\documentclass{article}

\usepackage{microtype}
\usepackage{graphicx}
\usepackage{subcaption}
\usepackage{booktabs} 

\usepackage{hyperref}

\usepackage[preprint]{icml2026}

\usepackage{amsmath}
\usepackage{amssymb}
\usepackage{mathtools}
\usepackage{amsthm}
\usepackage{booktabs}
\usepackage{multirow}
\usepackage{xcolor}
\usepackage[table]{xcolor}

\definecolor{highlightgray}{gray}{0.92}

\usepackage{xspace}
\newcommand{\ours}{\textsc{GenArena}\xspace}

\usepackage[capitalize,noabbrev]{cleveref}

\theoremstyle{plain}

\theoremstyle{definition}

\theoremstyle{remark}

\usepackage[textsize=tiny]{todonotes}

\usepackage[most]{tcolorbox}
\tcbuselibrary{listings}
\newtcblisting{promptbox}{
  breakable,
  listing only,
  title=PROMPT,
  fonttitle=\bfseries\color{black}, 
  coltitle=white,
  colbacktitle=white,               
  colframe=black,                   
  colback=white,                    
  boxrule=0.8pt,
  listing options={
    basicstyle=\ttfamily\small,
    breaklines=true,
    columns=fullflexible,
    numbers=none,
    backgroundcolor=\color{white}   
  }
}

\icmltitlerunning{GenArena: How Can We Achieve Human-Aligned Evaluation for Visual Generation Tasks?}

\begin{document}

\twocolumn[
  \icmltitle{
  GenArena: How Can We Achieve Human-Aligned Evaluation for Visual Generation Tasks?
}



  \icmlsetsymbol{equal}{*}

  \begin{icmlauthorlist}
    \icmlauthor{Ruihang Li}{ustc,sii,tx}
    \icmlauthor{Leigang Qu}{nus}
    \icmlauthor{Jingxu Zhang}{ustc}
    \icmlauthor{Dongnan Gui}{ustc} \\
    \icmlauthor{Mengde Xu}{tx}
    \icmlauthor{Xiaosong Zhang}{tx}
    \icmlauthor{Han Hu}{tx}
    \icmlauthor{Wenjie Wang}{ustc}
    \icmlauthor{Jiaqi Wang}{sii}
  \end{icmlauthorlist}

  \icmlaffiliation{ustc}{University of Science and Technology of China}
  \icmlaffiliation{sii}{Shanghai Innovation Institute}
  \icmlaffiliation{tx}{Tencent}
  \icmlaffiliation{nus}{National University of Singapore}

  \icmlcorrespondingauthor{Wenjie Wang}{wenjiewang@ustc.edu.cn}
  \icmlcorrespondingauthor{Jiaqi Wang}{wangjiaqi@sii.edu.cn}

  \icmlkeywords{}

\vspace{6pt}
\begin{center}
    \small
    \begin{tabular}{rl}
      \raisebox{-2pt}{\includegraphics[height=1.05em]{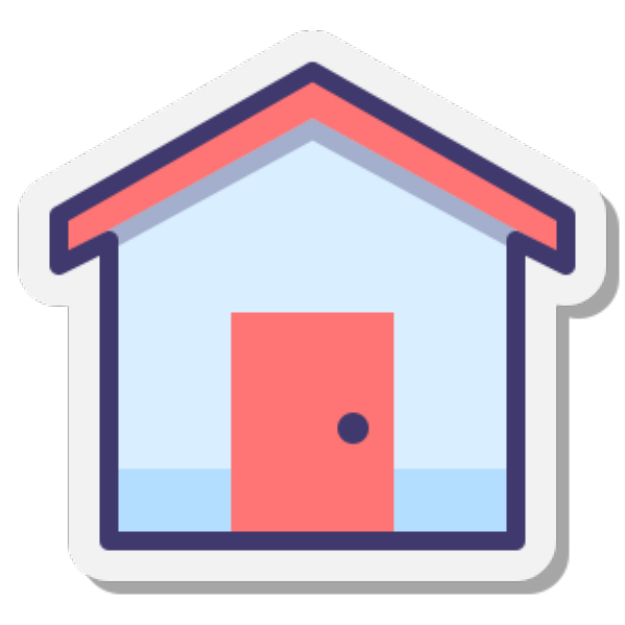}} & \url{https://genarena.github.io} \\
      \raisebox{-2pt}{\includegraphics[height=1.05em]{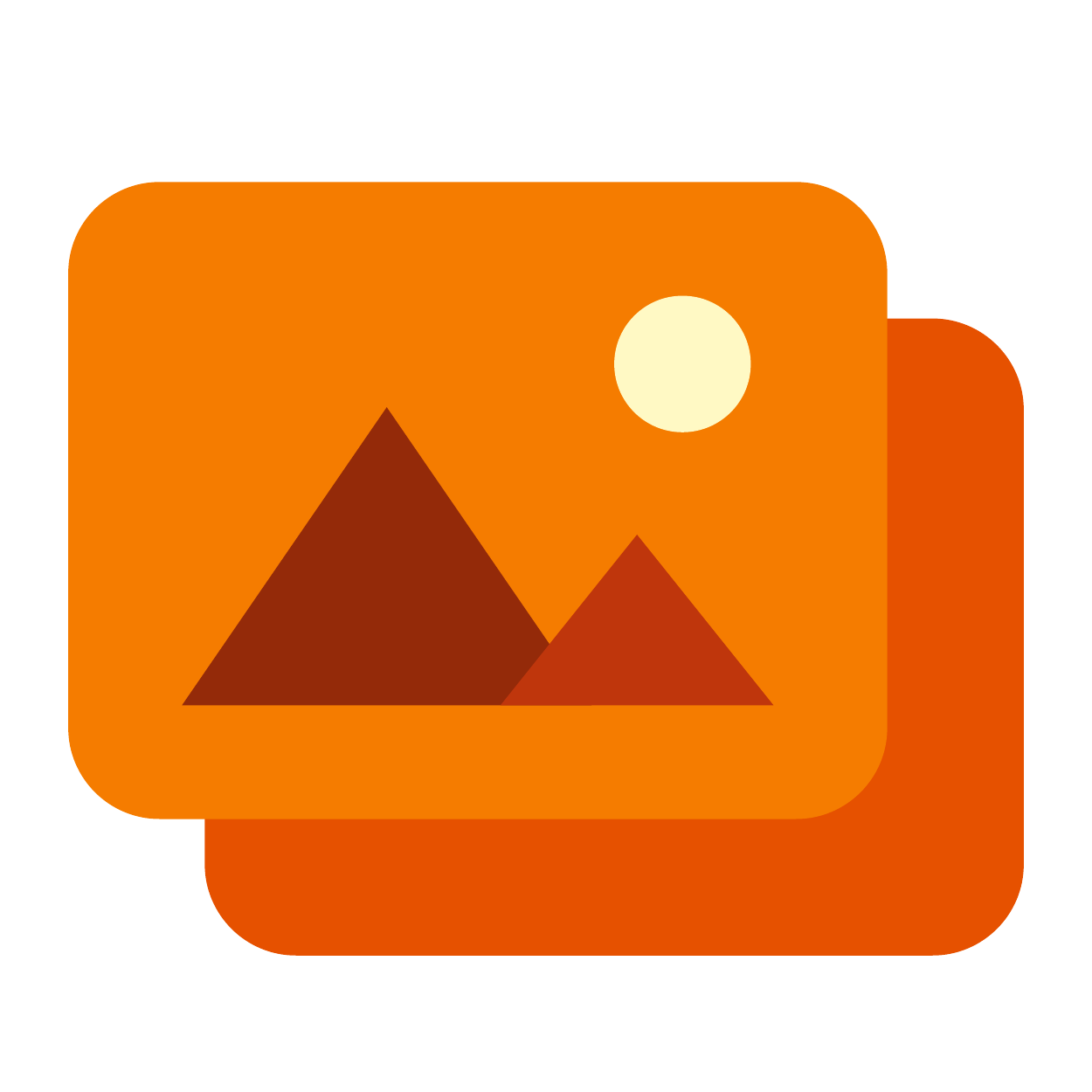}} & \url{https://huggingface.co/spaces/genarena/leaderboard} \\
      \raisebox{-2pt}{\includegraphics[height=1.05em]{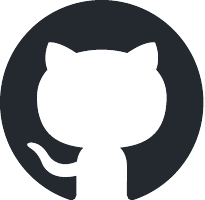}} & \url{https://github.com/ruihanglix/genarena} \\
      \raisebox{-2pt}{\includegraphics[height=1.05em]{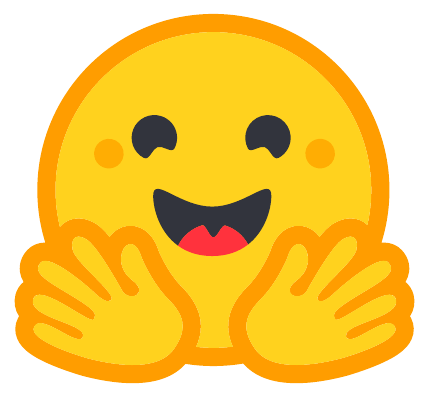}} & \url{https://huggingface.co/datasets/rhli/genarena}
    \end{tabular}
    \vspace{-6pt}
  \end{center}

  \vskip 0.3in
]



\printAffiliationsAndNotice{}  

\begin{abstract}
The rapid advancement of visual generation models has outpaced traditional evaluation approaches, necessitating the adoption of Vision-Language Models as surrogate judges. In this work, we systematically investigate the reliability of the prevailing absolute pointwise scoring standard, across a wide spectrum of visual generation tasks. Our analysis reveals that this paradigm is limited due to stochastic inconsistency and poor alignment with human perception. To resolve these limitations, we introduce \textbf{\ours}, a unified evaluation framework that leverages a \textit{pairwise comparison} paradigm to ensure stable and human-aligned evaluation. Crucially, our experiments uncover a transformative finding that simply adopting this pairwise protocol enables off-the-shelf open-source models to outperform top-tier proprietary models. Notably, our method boosts evaluation accuracy by over 20\% and achieves a Spearman correlation of 0.86 with the authoritative LMArena leaderboard, drastically surpassing the 0.36 correlation of pointwise methods. Based on \ours, we benchmark state-of-the-art visual generation models across diverse tasks, providing the community with a rigorous and automated evaluation standard for visual generation.
\end{abstract}

\section{Introduction}

\begin{figure*}[t] 
    \centering
    \includegraphics[width=\textwidth]{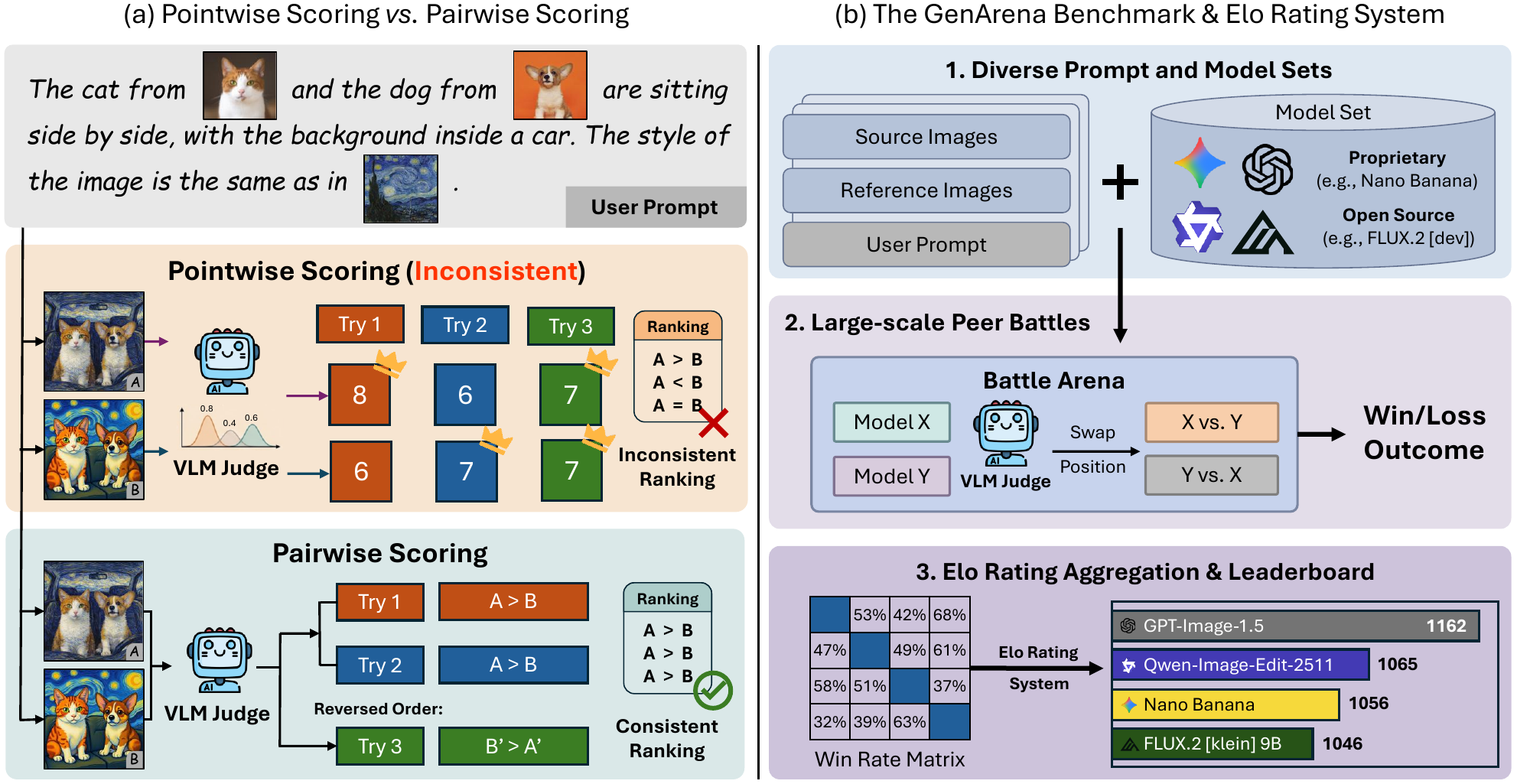}
\caption{
  \textbf{Schemetric illustration of the comparison between pointwise and pairwise scoring, the \ours Benchmark, and the Elo Rating System. }
  \textbf{(a)} Current benchmarks rely on absolute pointwise scoring, which suffers from \textit{self-consistency collapse}. As shown, stochastic fluctuations in VLM outputs result in volatile rankings (e.g., $A>B$ in Try 1, but $B>A$ in Try 2) for the same input. In contrast, pairwise comparison yields consistent and robust preferences.
  \textbf{(b)} \ours builds upon this stable pairwise paradigm. We curate a diverse set of prompts (including multi-reference generation tasks) and conduct large-scale peer battles using VLMs as judges. These pairwise outcomes are aggregated via the Elo Rating System to produce an accurate and reproducible model leaderboard.
  }
    \label{fig:method}
    \vspace{-5mm}
\end{figure*}

Recent advancements in diffusion models~\cite{ho2020denoising,song2020denoising,ldm,podell2023sdxl,blattmann2023stable,esser2024scaling} and unified multimodal models~\cite{team2024chameleon,wang2024emu3,pan2025transfer,bagel,xomni} have shifted the frontier of visual generation from basic text-to-image synthesis to complex tasks requiring multi-input reasoning, such as image editing~\cite{liu2025step1x,team2025longcat} and composition~\cite{labs2025flux1kontextflowmatching,wu2025qwen,flux-2-2025}. As these systems reach unprecedented levels of capability, establishing a rigorous and reliable evaluation standard becomes increasingly critical yet challenging. To fundamentally assess the efficacy of an evaluation protocol in this evolving landscape, an ideal judge should satisfy three core requirements: (1) \textit{Human Alignment}: exhibiting strong correlation with human perceptual rankings~\cite{zheng2023judging}; (2) \textit{Self-Consistency}: producing stable judgments for identical samples across independent trials~\cite{li2024llms}; and (3) \textit{Discriminability}: effectively distinguishing subtle differences among state-of-the-art models. Satisfying these rigorous standards is critical for visual generation, particularly for \textit{image editing and composition}, where correctness hinges on accurate prompt following and localized visual consistency that are hard to evaluate automatically.

However, satisfying these requirements remains challenging for traditional automated metrics. Widely used metrics such as FID~\cite{heusel2017gans,stein2023exposing,jayasumana2024rethinking} and CLIP Score~\cite{hessel2021clipscore} often fail to evaluate fine-grained semantic alignment and aesthetics~\cite{stein2023exposing,kynkaanniemi2023role,jayasumana2024rethinking}, struggling to capture the nuances required by high-fidelity generation. While human evaluation serves as the \textit{gold} standard, it incurs prohibitive costs and lacks scalability. Recently, in pursuit of a reliable yet automated alternative, the research community has pivoted towards the \textit{VLM-as-a-judge} paradigm~\cite{ku2024viescore,lin2024evaluating,imgedit,liu2025step1x}. By leveraging the general-purpose visual reasoning capabilities of Vision-Language Models (VLMs)~\cite{google2025gemini3flash,bai2025qwen3vltechnicalreport}, this paradigm promises a scalable path to automate evaluation while approximating human-like perception.

Within this paradigm, existing evaluation practices typically adopt either \textit{pointwise} scoring (assigning absolute scalar scores)~\cite{ku2024viescore,lin2024evaluating,imgedit,liu2025step1x} or \textit{pairwise} comparison (assessing relative quality)~\cite{wang2025pref,hu2025multimodal}. Despite their rapid adoption, the optimal methodology remains underexplored. First, regarding the scoring protocol, a fundamental question arises: \textit{1) Does the prevailing pointwise approach truly fulfill the core criteria of alignment and consistency, or does the pairwise paradigm offer a more robust alternative?} Second, regarding model accessibility, current high-quality evaluation relies predominantly on top-tier proprietary models~\cite{ku2024viescore,liu2025step1x,imgedit} to improve judgmental reliability, while open-source alternatives often lag behind. Bridging this gap typically necessitates fine-tuning on massive and costly human preference datasets~\cite{han2024evalmuse,ma2025hpsv3}, requiring continuous updates as visual generative models~\cite{wang2025unified,wu2025editreward,luo2025editscore} rapidly evolve. This raises further critical questions: \textit{2) Is reliable evaluation limited to proprietary models, or can open-source VLMs achieve state-of-the-art performance?} Furthermore, \textit{3) is such resource-heavy fine-tuning indispensable, or can off-the-shelf models serve as effective judges directly?}

To answer question 1), we conducted a comprehensive study spanning image generation, editing, and video generation. Our investigation reveals inherent limitations in the pointwise paradigm~\cite{liu2025step1x, imgedit}. As evidenced in Table~\ref{tab:pairwise_accurate}, Table~\ref{tab:ks-alpha}, and Figure~\ref{fig:pointwise_dist}, pointwise scoring suffers from poor human alignment and \textit{self-consistency collapse}, often failing to distinguish subtle
differences among models. We attribute this instability to cognitive limitations: maintaining a consistent absolute pointwise grading rubric is inherently prone to variance~\cite{ariely1998predictably}. In contrast, our analysis identifies the \textit{pairwise} paradigm—a strategy largely undervalued in current visual generation benchmarks—as the superior alternative. By simplifying evaluation into a robust binary choice~\cite{hu2025multimodal}, relative comparison effectively mitigates variance and restores alignment. 
For questions 2) and 3), experiments in Table~\ref{tab:pairwise_accurate} demonstrate that current open-source VLMs~\cite{bai2025qwen3vltechnicalreport,wang2025internvl3,vteam2025glm45vglm41vthinkingversatilemultimodal} possess strong discriminative capabilities that were previously obscured by the pointwise format. We find that adopting the pairwise protocol effectively unlocks this potential, enabling off-the-shelf open-source models~\cite{bai2025qwen3vltechnicalreport,vteam2025glm45vglm41vthinkingversatilemultimodal,wang2025internvl3} to achieve state-of-the-art accuracy without any parameter updates, surpassing even top-tier proprietary models using pointwise scoring.

\begin{table*}[t]
\centering
\caption{ \textbf{Pairwise scoring universally outperforms pointwise scoring across a wide spectrum of visual generation tasks.} We report the binary classification accuracy with human preferences on GenAI-Bench (``GAI" for short)~\cite{li2024genai}, EditScore-Bench~\cite{luo2025editscore}, and VideoGen-RewardBench~\cite{liu2025improving}. The results demonstrate that simply shifting to a pairwise protocol (rows with \textcolor{gray}{gray} background) enables off-the-shelf open-source VLMs to achieve state-of-the-art accuracy without any parameter updates, consistently surpassing specialized reward models and proprietary systems like GPT-5 that rely on the pointwise paradigm. \textbf{Bold} indicates the best result, and \underline{underline} denotes the second best.}
\label{tab:pairwise_accurate}
\resizebox{0.85\textwidth}{!}{
\begin{tabular}{l c c c c c}
\toprule
\multirow{2}{*}{\textbf{Model}} & \textbf{Image Generation} & \multicolumn{2}{c}{\textbf{Image Editing}} & \multicolumn{2}{c}{\textbf{Video Generation}} \\
\cmidrule(lr){2-2} \cmidrule(lr){3-4} \cmidrule(lr){5-6}
& \textbf{GenAI-Bench} & \textbf{GAI} & \textbf{EditScore} & \textbf{GAI} & \textbf{VideoGen-Reward} \\
\midrule
\multicolumn{6}{l}{\textit{Proprietary VLMs}} \\
GPT-4.1                        & -- & -- & 70.5 & -- & -- \\
GPT-5                          & -- & -- & 75.5 & -- & -- \\
Gemini-2.5 Pro                 & -- & -- & 72.2 & -- & -- \\
\midrule
\multicolumn{6}{l}{\textit{Finetuned Models}} \\
VisionReward & \underline{66.4} & -- & -- & 73.1 & 68.2 \\
Qwen2.5-VL-7B                  & -- & -- & 43.2 & -- & -- \\
\quad EditScore-7B             & -- & -- & 65.9 & -- & -- \\
Qwen2.5-VL-72B                 & -- & -- & 62.1 & -- & -- \\
\quad EditScore-72B            & -- & -- & 70.3 & -- & -- \\
Qwen3-VL 8B Instruct           &  &  &  &  & \\
\quad EditScore-Qwen3-8B             & -- & -- & 69.0 & -- & -- \\
\quad UnifiedReward-Qwen3-VL-8B & 64.2 & 81.5 & 75.0 & 69.1 & \underline{73.6} \\
\rowcolor{highlightgray}
\quad \quad + Pairwise               & \textbf{67.0} & 82.5 & 73.3 & \textbf{78.6} & \textbf{78.8} \\
\midrule
\multicolumn{6}{l}{\textit{Open Source VLMs}} \\
Qwen3-VL 8B Instruct           & 49.1 & 73.4 & 58.3 & 62.0 & 57.0 \\
\rowcolor{highlightgray}
\quad + Pairwise               & 60.5 & \textbf{83.9} & \underline{83.7} & 73.3 & 61.5 \\
GLM 4.6V Flash (9B)              & 48.2 & 73.2 & 68.3 & 63.0 & -- \\
\rowcolor{highlightgray}
\quad + Pairwise               & 54.7 & 81.3 & \textbf{87.2} & \underline{76.6} & -- \\
InternVL 3.5 8B                & 50.7 & 66.4 & 53.4 & -- & -- \\
\rowcolor{highlightgray}
\quad + Pairwise               & 61.9 & \underline{83.1} & 75.0 & -- & -- \\
\bottomrule
\end{tabular}
}
\vspace{-2mm}
\end{table*}

Building upon these findings, we introduce \textbf{\ours}, a standardized evaluation framework depicted in Figure~\ref{fig:method} (b). Instead of relying on unstable pointwise scores, \ours constructs dynamic leaderboards by aggregating the outcomes of massive pairwise comparisons via the Elo Rating System~\cite{elo1966uscf}. By leveraging the inherent multi-image reasoning capabilities of VLMs~\cite{bai2025qwen3vltechnicalreport,vteam2025glm45vglm41vthinkingversatilemultimodal,wang2025internvl3} to adjudicate peer battles, we ensure that model rankings are grounded in consistent comparative criteria. While this framework is universally applicable to visual generation, we specifically target \textit{image editing and composition}, where the disparity between advancing model capabilities~\cite{google2025gemini25flashimage,googledeepmind2025geminiimagepro,gptimage1,openai2025gptimage15} and lagging evaluation protocols are most pronounced. Specifically, we curate a comprehensive evaluation suite comprising 6,086 high-quality user prompts spanning three categories: basic editing, reasoning-intensive editing, and multi-reference composition. Based on the suite, we conduct an extensive evaluation, as shown in Table~\ref{tab:pairwise_accurate}. It shows that switching to \textit{pairwise} scoring paradigm enables off-the-shelf open-source VLMs~\cite{bai2025qwen3vltechnicalreport,vteam2025glm45vglm41vthinkingversatilemultimodal,wang2025internvl3} to achieve a \textbf{+20\%} accuracy boost, surpassing even top-tier proprietary models. Crucially, our framework yields a Spearman correlation of \textbf{0.86} with human preference, drastically outperforming the 0.36 correlation of pointwise baselines (Table~\ref{tab:elo_more_align}).

\begin{table*}[t]
\centering
\caption{
\textbf{\ours Leaderboard.} Elo-based rankings of a variety of SoTA models across Basic, Reasoning, and Multi-Reference editing tasks. Rankings are established via pairwise battles judged by \textit{Qwen3-VL-32B Instruct FP8} and aggregated into Elo scores. The results demonstrate that while basic editing capabilities are converging, significant disparities remain in reasoning and multi-reference tasks, distinguishing frontier proprietary models from current open-source alternatives. The rightmost column shows rankings from the LMArena~\cite{lmarena} leaderboard (Jan 16, 2026 version).
}
\label{tab:bench_models}
\rowcolors{3}{white}{highlightgray}
\begin{tabular}{lccccccc}
\toprule
\multirow{2}{*}{\textbf{Models}} & \multicolumn{2}{c}{Basic} & \multicolumn{2}{c}{Reasoning} & \multicolumn{2}{c}{MultiRef} & \multirow{2}{*}{\color{gray}LMArena} \\
\cmidrule(lr){2-3} \cmidrule(lr){4-5} \cmidrule(lr){6-7}
 & Elo & Rank & Elo & Rank & Elo & Rank & \\
\midrule
GPT Image 1.5 [High]~\cite{openai2025gptimage15} & 1162 & \#1 & 1204 & \#1 & 1259 & \#1 & \color{gray}\#1 \\
Qwen-Image-Edit-2511~\cite{wu2025qwen} & 1065 & \#2 & 1005 & \#4 & 793 & \#7 & \color{gray}\#3 \\
Nano Banana~\cite{google2025gemini25flashimage} & 1056 & \#3 & 1130 & \#2 & 1048 & \#3 & \color{gray}\#2 \\
Flux.2 [klein] 9B~\cite{blackforestlabs2026flux2klein} & 1046 & \#4 & 962 & \#6 & 1018 & \#4 & \color{gray}\#5 \\
LongCat-Image-Edit~\cite{team2025longcat} & 1037 & \#5 & 944 & \#8 & -- & -- & \color{gray}-- \\
Qwen-Image-Edit-2509~\cite{wu2025qwen} & 1020 & \#6 & 962 & \#7 & 705 & \#10 & \color{gray}-- \\
GPT Image 1 [High]~\cite{gptimage1} & 1004 & \#7 & 1095 & \#3 & 1066 & \#2 & \color{gray}\#9 \\
Flux.2 [dev]~\cite{flux-2-2025} & 997 & \#8 & 968 & \#5 & 948 & \#6 & \color{gray}\#4 \\
Flux.2 [klein] 4B~\cite{blackforestlabs2026flux2klein} & 987 & \#9 & 928 & \#9 & 967 & \#5 & \color{gray}\#7 \\
Qwen-Image-Edit~\cite{wu2025qwen} & 979 & \#10 & 920 & \#10 & -- & -- & \color{gray}\#6 \\
Flux.1 Kontext [dev]~\cite{labs2025flux} & 860 & \#11 & 849 & \#12 & 713 & \#9 & \color{gray}\#8 \\
Bagel~\cite{bagel} & 773 & \#12 & 823 & \#13 & 649 & \#11 & \color{gray}\#10 \\
Step1X-Edit~\cite{liu2025step1x} & 739 & \#13 & 744 & \#14 & -- & -- & \color{gray}\#11 \\
DreamOmni2~\cite{xia2025dreamomni2} & 718 & \#14 & 858 & \#11 & 777 & \#8 & \color{gray}-- \\
\midrule
\rowcolor{white}
\textbf{Corr. w/ LMArena} & \multicolumn{2}{c}{0.87} & \multicolumn{2}{c}{0.80} & \multicolumn{2}{c}{0.50} & \color{gray}-- \\
\bottomrule
\end{tabular}
\end{table*}

The contributions of this work are summarized as follows:

\begin{enumerate}
    \item We identify the critical flaws of absolute pointwise scoring, specifically \textit{self-consistency collapse} and \textit{poor human alignment}. In contrast, we demonstrate that shifting to a pairwise paradigm serves as a more robust and accurate alternative for visual evaluation.
    
    \item We propose \textbf{\ours}, an Elo-based benchmark for visual generation tasks based on the pairwise scoring mechanism. By unifying assessments ranging from basic editing to complex multi-reference generation, \ours enables rigorous, consistent, and fair model comparison.
    
    \item We validate that our benchmark achieves superior correlation with human preference compared to existing methods. Our experiments verify that such a pairwise scoring approach unlocks the potential of open-sourced VLMs for visual generation evaluation, establishing a reproducible and highly accurate standard for the research community.
\end{enumerate}

\section{Related Work}
\paragraph{LLM-as-a-Judge.} The capacity of Large Language Models (LLMs) to emulate human reasoning and evaluate specific inputs against predefined criteria has established the \textit{LLM-as-a-Judge} paradigm~\cite{zheng2023judging,chen2024mllm,li2024llms,li2025generation}. This approach has been widely adopted in visual generation~\cite{ku2024viescore,imgedit,liu2025step1x,han2025unireditbench}, demonstrating high alignment with human preferences, and has recently been extended to Generative Reward Models to provide precise feedback signals for reinforcement learning~\cite{liu2025inference,wang2025unified}. Nevertheless, these judges remain prone to vulnerabilities that introduce confounding factors into the evaluation, such as position bias~\cite{ye2024justice,shi2024judging,wang2024large,thakur2025judging}, self-enhancement bias~\cite{tan2024blinded,ye2024justice}, overconfidence~\cite{Gu2024ASO}, and susceptibility to jailbreaking attacks~\cite{raina2024llm,zheng2024cheating,zhao2025one}. In this work, we identify that stochastic inconsistency across repeated evaluations is a critical issue within this paradigm, a problem significantly exacerbated by standard pointwise scoring methods.

\paragraph{Visual Generation Benchmarks.} 
Benchmarks for visual generative models have expanded rapidly across text-to-image generation~\cite{ghosh2023geneval,huang2023t2i,sun2025t2i,niu2025wise}, text-to-video generation~\cite{liu2023fetv,liu2024evalcrafter,huang2024vbench}, and image editing tasks~\cite{brooks2023instructpix2pix,zhang2023magicbrush,sheynin2024emu,liu2025step1x,imgedit,wu2025kris,han2025unireditbench}. While methodologies have evolved from low-level statistics (e.g., CLIP, SSIM)~\cite{radford2021learning,korhonen2012peak,wang2004image} and detection-based verification~\cite{ghosh2023geneval,huang2023t2i} to recent ``VLM-as-a-Judge'' paradigms~\cite{imgedit,ku2024viescore}, a fundamental limitation persists. Whether assessing compositional constraints~\cite{ghosh2023geneval,huang2023t2i}, temporal consistency~\cite{huang2024vbench,liu2024evalcrafter}, or instruction following~\cite{liu2025step1x}, existing protocols predominantly rely on \textit{absolute pointwise scoring}. As noted in recent critiques~\cite{stein2023exposing,kynkaanniemi2023role}, these scalar metrics suffer from stochastic inconsistency and the ambiguity of absolute scales, often yielding rankings that diverge from human perception. To address these shortcomings, we propose a scalable \textit{pairwise} comparison framework~\cite{thurstone2017law} grounded in the Elo rating system~\cite{elo1966uscf} to ensure robust and reproducible evaluation.

\paragraph{Elo Rating System in AI Evaluation.} The practice of ranking model capabilities via Elo scores derived from pairwise comparisons has become a standard in the evaluation of both LLMs~\cite{team2025kimi,xai2025grok,google2025gemini3flash} and visual generative models~\cite{labs2025flux1kontextflowmatching, wu2025qwen, cao2025hunyuanimage}. Public benchmarks such as LMArena~\cite{lmarena}, Artificial Analysis Arena~\cite{aaarena} and GenAI-Arena~\cite{genaiarena} leverage crowdsourced voting, where users blindly select the superior response from two models, to calculate win rates and subsequent Elo ratings. Subsequent works~\cite{zhao2025auto,luo2025videoautoarena} further automated this process by utilizing LLMs to generate queries and evaluate responses, thereby reducing the dependency on manual labor. To the best of our knowledge, this work represents the first integration of an automated Elo rating system into the domain of visual generation and editing. We employ the Elo mechanism to effectively harness the superior comparative reasoning capabilities of VLMs, establishing a benchmark that is both robust and accurate.

\section{Revisiting the VLM-as-a-judge Paradigm}
\label{sec:protocol_analysis}
While VLMs are increasingly used as surrogate judges, the standard practice of absolute pointwise scoring remains unexamined. In this section, we conduct a systematic study to explore the reliability of this paradigm. By comparing it with a pairwise approach across diverse tasks, we uncover its significant limitations in both accuracy (Sec.~\ref{sec:pairwise_accuracy}) and stability (Sec.~\ref{sec:ks-alpha}). Consequently, our results establish that the pairwise protocol offers a far more robust and accurate evaluation standard, providing the empirical cornerstone for \ours.

\subsection{Pairwise scoring is more accurate than its pointwise alternatives}
\label{sec:pairwise_accuracy}

In this section, we quantitatively validate the superiority of the pairwise paradigm over traditional absolute pointwise scoring across a wide spectrum of visual generation tasks. To isolate the impact of the scoring protocol from model capability, we evaluate a diverse set of open-source VLMs~\cite{vteam2025glm45vglm41vthinkingversatilemultimodal,wang2025internvl3,bai2025qwen3vltechnicalreport} under both pointwise and pairwise settings. Our evaluation leverages high-quality human-annotated preference datasets spanning three distinct domains. For image generation, we utilize GenAI-Bench~\cite{li2024genai}. For image editing, we employ GenAI-Bench alongside EditScore-Bench~\cite{luo2025editscore}. For video generation, we use GenAI-Bench~\cite{li2024genai} and VideoGen-RewardBench~\cite{liu2025improving}. We report the pairwise accuracy, defined as the agreement rate with ground-truth human choices, as our primary metric to assess the reliability of each judge.

The results summarized in Table~\ref{tab:pairwise_accurate} reveal a systematic performance disparity where pairwise scoring consistently outperforms pointwise scoring across image generation, editing, and video tasks. We observe that the evaluation protocol itself serves as a critical performance lever. Simply switching from a pointwise to a pairwise paradigm triggers an immediate and substantial accuracy boost for identical models without any parameter updates. For instance, Qwen3-VL 8B Instruct achieves a remarkable leap across different modalities, improving accuracy from 49.1\% to 60.5\% on GenAI-Bench~\cite{li2024genai} of image generation, rising from 58.3\% to 83.7\% on EditScore-Bench~\cite{luo2025editscore} of image editing, and advancing from 57.0\% to 61.5\% on VideoGen-RewardBench~\cite{liu2025improving} of video generation. Similar consistent gains across different VLM judges confirm that \textit{the pairwise paradigm effectively mitigates calibration noise, thereby unlocking the latent discriminative power of VLMs}.

Furthermore, this methodological shift \textit{enables general-purpose models to surpass specialized systems that are significantly larger or explicitly finetuned for evaluation}. As evidenced in Table~\ref{tab:pairwise_accurate}, the off-the-shelf Qwen3-VL 8B Instruct utilizing pairwise scoring outperforms the specialized EditScore-72B~\cite{luo2025editscore} reward model in editing and surpasses VisionReward~\cite{xu2024visionreward} in video tasks, despite both baselines relying on pointwise scoring. Moreover, this robust protocol allows open-source VLMs to challenge top-tier proprietary models. GLM-4V-Flash with our pairwise protocol attains 87.2\% accuracy on EditScore-Bench~\cite{luo2025editscore}, significantly exceeding GPT-5 at 75.5\%. In light of its superior accuracy, efficiency and widespread community adoption, we designate Qwen3-VL 8B Instruct~\cite{bai2025qwen3vltechnicalreport} as the default judge model for subsequent experiments unless otherwise specified.

\begin{table}[thbp]
  \centering
  \caption{\textbf{Comparison of Judge Self-Consistency (Krippendorff’s $\alpha$).} We report the $\alpha$ values calculated over 5 independent inference runs using \textit{Qwen3-VL 8B Instruct} as the judge. The pairwise paradigm demonstrates superior internal stability compared to the pointwise baseline across both human-annotated datasets and prompts from commonly-used image editing benchmarks. \textbf{Bold} denote the best results.}  
  \label{tab:ks-alpha}
\resizebox{0.9\linewidth}{!}{
  \begin{tabular}{lcc}
    \toprule
    & \textbf{Pointwise} & \textbf{Pairwise} \\
    \midrule
    \textit{Reward Model Benchmarks} & & \\
    GenAI-Bench & 0.7256 & \textbf{0.8628} \\
    EditScore-Bench & 0.5753 & \textbf{0.7087} \\
    \midrule
    \textit{Image Editing Benchmarks} & & \\
    ImgEdit & 0.5707 & \textbf{0.7040} \\
    GEdit-Bench & 0.5169 & \textbf{0.6553} \\
    \bottomrule
  \end{tabular}
}
  \vspace{-3mm}
\end{table}

\subsection{Pairwise scoring is more consistent than its pointwise alternatives}
\label{sec:ks-alpha}

\begin{table*}[t]
\centering
\caption{\textbf{Elo rating aligns better with LMArena than pointwise scoring.} We compare the ranking alignment of absolute pointwise scores and our pairwise Elo ratings on the widely-used GEdit-Bench-EN~\cite{liu2025step1x} prompt set against human preference represented by LMArena~\cite{lmarena}. For the pointwise baseline, we source scores from~\citet{yang2025wemmuenhancedbridgingvisionlanguage} for Nano Banana,~\citet{li2025uniworld} for Flux.1 Kontext [dev], and~\citet{wu2025qwen} for other models, with Flux.2 [dev]~\cite{flux-2-2025} evaluated by us. Compared with pointwise scoring yielding a low correlation of $\rho=0.36$, our pairwise approach achieves more robust alignment with human judgment, reaching $\rho=0.86$.}
\label{tab:elo_more_align}
\begin{tabular}{l cc cc c}
\toprule
\multirow{2}{*}{\textbf{Models}} & \multicolumn{2}{c}{GEdit-EN (Point)} & \multicolumn{2}{c}{\textbf{GEdit-EN (Elo)}} & \multirow{2}{*}{\textcolor{gray}{LMArena}} \\
\cmidrule(lr){2-3} \cmidrule(lr){4-5}
 & Score & Rank & Score & Rank & \\
\midrule
Nano Banana~\cite{google2025gemini25flashimage}       & 7.10 & 4 & 1062 & 1 & \textcolor{gray}{1325} \\
Flux.2 [dev]~\cite{flux-2-2025}                       & 7.24 & 3 & 1053 & 2 & \textcolor{gray}{1250} \\
Qwen-Image-Edit~\cite{wu2025qwen}                     & 7.56 & 1 & 992 & 4 & \textcolor{gray}{1231} \\
Flux.1 Kontext [dev]~\cite{labs2025flux}              & 6.00 & 7 & 964  & 5 & \textcolor{gray}{1166} \\
GPT Image 1 [High]~\cite{gptimage1}                   & 7.53 & 2 & 1034 & 3 & \textcolor{gray}{1155} \\
Bagel~\cite{bagel}                                    & 6.52 & 6 & 890  & 7 & \textcolor{gray}{1042} \\
Step1X-Edit~\cite{liu2025step1x}                      & 6.70 & 5 & 938  & 6 & \textcolor{gray}{1014} \\
\midrule
Corr. w/ LMArena~\cite{lmarena} & \multicolumn{2}{c}{0.36} & \multicolumn{2}{c}{\textbf{0.86}} &  \\
\bottomrule
\end{tabular}
\vspace{-3mm}
\end{table*}

Beyond alignment with human preference, a robust benchmark demands \textit{internal stability}, ensuring that stochastic fluctuations during decoding do not alter rankings. We quantify this self-consistency using \textit{Krippendorff’s Alpha} ($\alpha$)~\cite{krippendorff2011computing}, treating independent inference runs with identical inputs as distinct raters. To enable direct comparison, we project pointwise scalar scores into a unified categorical preference space, utilizing a nominal difference metric that only penalizes fluctuations severe enough to flip the decision boundary (see Appendix~\ref{app:consistency} for formal definitions).

We evaluate consistency across both human-annotated datasets~\cite{li2024genai,luo2025editscore} and large-scale model matchups~\cite{imgedit,liu2025step1x}. As detailed in Table~\ref{tab:ks-alpha}, the absolute pointwise scoring paradigm suffers from \textit{self-consistency collapse}, with $\alpha$ dropping as low as 0.5169 on GEdit-Bench. In stark contrast, our pairwise framework significantly mitigates this noise, maintaining high stability across all benchmarks (e.g., $\alpha=0.8628$ on GenAI-Bench). These findings confirm that transitioning to comparative reasoning effectively rectifies the inherent volatility of pointwise metrics.

These findings suggest that reliable automated evaluation benefits more from pairwise preferences rather than pointwise scores. Motivated by these findings, we introduce \ours, a scalable framework to aggregate pairwise outcomes into reproducible leaderboards via Elo ratings.

\section{GenArena}
\label{sec:genarena}

\subsection{Overview}
\label{sec:method_overview}

In this section, we introduce \textbf{\ours}, a standardized evaluation framework designed to overcome the ``self-consistency collapse'' of pointwise scoring by shifting to a robust comparative paradigm. As illustrated in Figure~\ref{fig:method}, \ours orchestrates a scalable tournament among generative models, emulating the dynamics of human preference aggregation in a fully automated, reproducible pipeline:

\textbf{Stage 1: Competitive Sampling.} We first curate a diverse instruction set $\mathcal{I}$ and sample outputs from a pool of candidate models $\mathcal{M} = \{M_1, \dots, M_K\}$, treating them as active competitors in a tournament rather than isolated subjects.

\textbf{Stage 2: Robust Pairwise Judging.} A VLM judge evaluates model pairs $(M_i, M_j)$. To mitigate the stochasticity and position bias inherent in LLMs, we implement a \textit{bi-directional consistency check} and strictly enforce a \textit{forced-choice} mechanism, only resolving those low-confidence (conflict) judgments into ties (see Sec.~\ref{sec:pairwise_scoring}).

\textbf{Stage 3: Global Elo Aggregation.} Discrete pairwise outcomes are transformed into a continuous leaderboard using the Bradley-Terry statistical model, ensuring a statistically rigorous ranking (see Sec.~\ref{sec:elo_ranking}).

\subsection{Benchmark Composition} 
\label{sec:benchmark_composition}

To ensure \ours serves as a holistic standard for challenging visual generation tasks, particularly image editing and multi-reference image composition, we curate a comprehensive evaluation suite consisting of \textbf{6,086} high-quality prompts (as detailed in Appendix~\ref{app:dataset_details}). We categorize prompts into three distinct capability dimensions to probe the frontiers of model performance:

\textbf{Basic Instruction Editing (1,948 prompts).} This subset evaluates fundamental instruction-following capabilities, including object modification, attribute alteration, and background changes. It aggregates filtered prompts from established datasets~\cite{imgedit,liu2025step1x,hu2025multimodal} to ensure broad coverage of standard editing scenarios.

\textbf{Reasoning-Intensive Editing (1,627 prompts).} Modern models should handle queries requiring complex logic, spatial reasoning, and world knowledge. We compile implicit and context-dependent instructions (e.g., ``make the image look like a scene from the 1920s'')~\cite{zhao2025envisioning,wu2025kris} that challenge the model to infer visual edits from abstract textual descriptions.

\textbf{Multi-Reference Composition (2,511 prompts).} To compensate for the lack of multi-reference generation in existing evaluations, this subset targets the integration of multiple visual concepts. We curate challenging prompts~\cite{wu2025omnigen2,xia2025dreamomni2,oshima2025multibanana,hu2025multimodal} requiring the coherent composition of multiple subject images, reflecting the complexity of real-world workflows like subject-driven generation~\cite{wu2025qwen,wu2025omnigen2,xia2025dreamomni2,flux-2-2025}.


\subsection{Automatic Pairwise Scoring}
\label{sec:pairwise_scoring}

To mitigate the calibration drift and inconsistency inherent in absolute pointwise scoring, we adopt a \textit{pairwise comparison paradigm}. In detail, a VLM judges a triplet $(I, O_A, O_B)$, comprising an instruction $I$ and two model outputs, i.e., $O_A$ and $O_B$, to determine the superior candidate. To ensure robustness against position bias and stochasticity~\cite{zheng2023judging}, we implement a \textit{Bi-directional Consistency Protocol} (detailed in Appendix~\ref{app:pairwise_details}). 

Besides, we enforce a \textit{forced-choice constraint}, prohibiting the judge from declaring ties during a single inference pass. Specifically, we query the judge twice with the image order swapped: $(O_A, O_B)$ and $(O_B, O_A)$. A reliable preference is recorded only if the judge model consistently selects the same image across both permutations (i.e., $A \succ B$ and $B \prec A$). Cases of conflicting preferences (e.g., the judge always selects the first position) are algorithmically resolved as a Tie ($S_{ij}=0.5$). This rigorous filtering mechanism ensures that only high-confidence, non-stochastic judgments contribute to the final ranking.

\subsection{Elo Ranking Aggregation}
\label{sec:elo_ranking}

To aggregate discrete pairwise outcomes into a continuous global ranking, we employ the \textbf{Elo rating system}~\cite{elo1966uscf} modeled via the Bradley-Terry (BT) framework~\cite{bradley1952rank}. We define the probability that model $i$ defeats model $j$ as a logistic function of the difference in their latent skill ratings $R_i$ and $R_j$:
\begin{equation}
    P(i \succ j) = \frac{1}{1 + 10^{(R_j - R_i) / \xi}},
\end{equation}
where $\xi=400$ is a scaling factor standard in Elo systems.

Given a dataset of pairwise comparisons, we estimate the optimal Elo scores $\mathbf{R} = \{R_1, \dots, R_N\}$ by performing Maximum Likelihood Estimation (MLE). Let $W_{ij}$ denote the total number of wins model $i$ achieves against model $j$. Following standard practices~\cite{lmarena, zhao2025auto}, we incorporate ties generated by our consistency protocol by allocating $0.5$ wins to both models. The optimal scores are obtained by maximizing the log-likelihood function:

\vspace{-7mm}
\begin{equation}
    \begin{split}
        \mathcal{L}(\mathbf{R}) &= \sum_{i \neq j} W_{ij} \ln P(i \succ j) \\
        &= \sum_{i \neq j} W_{ij} \ln \left( \frac{1}{1 + 10^{(R_j - R_i) / 400}} \right).
    \end{split}
\end{equation}
\vspace{-6mm}

This formulation allows us to solve for the global ranking that best explains the observed pairwise preferences, providing a statistically grounded leaderboard resilient to individual judgments.

\section{Evaluation and Benchmarking}
\label{sec:experiments}

In this section, we validate the effectiveness of \ours by demonstrating its superior human alignment compared to pointwise baselines (Sec.~\ref{sec:elo_more_align}). Furthermore, we benchmark various state-of-the-art image editing models (Sec.~\ref{sec:benchmark}) and ablate our design choices (Sec.~\ref{sec:method_validation}).

\subsection{Elo ranking aligns more with human preferences}
\label{sec:elo_more_align}

To study the alignment with human preference, we conduct a comparative study using the prompt set from GEdit-Bench~\cite{liu2025step1x} across the models listed in Table~\ref{tab:elo_more_align}. We assess the candidate models under two protocols: the standard absolute pointwise scoring (Point) utilizing GPT-4.1, and our proposed pairwise Elo rating system (Elo) employing Qwen3-VL 8B Instruct~\cite{bai2025qwen3vltechnicalreport} as the judge. We then benchmark the resulting rankings against the authoritative LMArena~\cite{lmarena}, which serves as the ground-truth proxy for crowdsourced human judgment.

As presented in Table~\ref{tab:elo_more_align}, the choice of evaluation paradigm significantly impacts ranking reliability. The absolute pointwise baseline struggles to reflect human consensus, yielding a low Spearman correlation~\cite{spearman1961proof} of 0.36 and notably misranking top-performing models like \textit{Nano Banana} by placing it the $4^{th}$ in GEdit-EN (Point) compared to $1^{st}$ in LMArena. In contrast, even when relying on a smaller open-source judge, the pairwise Elo approach effectively mitigates this noise, recovering a hierarchy that closely mirrors the ground truth. This methodological shift drives a substantial improvement in correlation to 0.86, demonstrating that converting the VLM's task from pointwise scoring to comparative ranking is the key factor in aligning automated metrics with human perception.

\subsection{Benchmarking existing generation models}
\label{sec:benchmark}
To establish a reliable leaderboard, we first determine the optimal evaluator based on the empirical evidence from Section~\ref{sec:judge_alignment}. Our experiments indicate that \textit{Qwen3-VL-32B Instruct FP8} yields the highest comprehensive agreement with human preference across the Basic, Reasoning, and MultiRef evaluation dimensions. Consequently, we adopt it as our designated judge to evaluate various state-of-the-art models. The assessment is conducted across three distinct tracks: \textit{Basic Editing} for standard instruction following and visual quality, \textit{Reasoning Editing} for complex spatial and logical constraints, and \textit{MultiRef} for composition involving multiple reference images. We aggregate the outcomes of pairwise battles into global rankings using the Elo rating system, as presented in Table~\ref{tab:bench_models}, with visualized results provided in Figure~\ref{fig:samples_of_models} and Figure~\ref{fig:samples_of_models_multiref}.

\begin{table}[t]
\centering
\caption{\textbf{Confusion Matrix of Model Predictions with Explicit Tie Option.} Despite a moderate overall accuracy, the model exhibits significant \textit{Laziness Bias}. In cases where humans identify a clear winner (A or B is better), the model fails to discriminate and defaults to ``Tie" in nearly 40\% of samples (highlighted with \underline{underline}).}
\label{tab:tie_ablation}
\resizebox{0.9\linewidth}{!}{
\begin{tabular}{l|ccc}
\toprule
\multirow{2}{*}{\textbf{Human Judgment}} & \multicolumn{3}{c}{\textbf{Model Prediction}} \\
\cmidrule(lr){2-4}
& \textbf{A $>$ B} & \textbf{B $>$ A} & \textbf{Tie (A$=$B)} \\
\midrule
\textbf{A $>$ B} & 55.8\% & 6.2\% & \underline{37.9\%} \\
\textbf{B $>$ A} & 6.8\% & 54.0\% & \underline{39.2\%} \\
\textbf{Tie} & 17.8\% & 16.8\% & 65.4\% \\
\bottomrule
\end{tabular}
}
\vspace{-6mm}
\end{table}

The experimental results present the model performance across Basic, Reasoning, and MultiRef categories, revealing that competency in standard visual editing does not necessarily translate to effectiveness in complex scenarios. While \textit{GPT Image 1.5 [High]} maintains a consistent lead, we observe notable divergences in other models. For instance, \textit{Qwen-Image-Edit-2511} excels in \textit{Basic} tasks but ranks lower in \textit{MultiRef}, whereas \textit{GPT Image 1 [High]} shows the opposite trend by performing significantly better in composition tasks. Quantitatively, our rankings achieve a high Spearman correlation of 0.87 with the LMArena leaderboard~\cite{lmarena}. Although the correlation is lower for \textit{MultiRef}, we attribute this discrepancy to the simpler prompt distribution in LMArena compared to our targeted complicated composition tasks. Qualitative results in Figure~\ref{fig:samples_of_models_multiref} further confirm that our judge remains highly accurate in evaluating these complex generation qualities.

\subsection{Validation of Design Choices}
\label{sec:method_validation}

In the following, we explore the design choices of the \ours framework. Specifically, we investigate two critical components that define the reliability of our pairwise judge: the strategic handling of tie scenarios to mitigate judge inertia, and the impact of the surrogate model's capacity on human alignment.

\subsubsection{Should we allow tie?}
\label{sec:allow-tie}

We analyze the impact of the scoring strategy by comparing an ``Explicit Tie'' option against our ``Forced Choice'' paradigm. We perform this ablation experiment on GenAI-Bench~\cite{li2024genai}, which includes \textit{tie} annotations by human. As detailed in Table~\ref{tab:tie_ablation}, allowing explicit ties introduces severe \textit{Laziness Bias}: the model defaults to a neutral judgment in nearly 40\% of cases where humans identify a clear winner ($A>B$ or $B>A$). This high False Neutral Rate suggests the model frequently uses the tie option to evade complex reasoning rather than reflecting true equivalence.

This bias significantly hurts the judge's performance. On the subset of discriminative pairs (where Ground Truth $\neq$ Tie), the Explicit Tie strategy achieves only 54.9\% accuracy due to these missed judgments. In contrast, the Forced Choice strategy compels the model to resolve ambiguity, boosting accuracy on the same subset to 83.9\%. The resulting \textbf{+29.0\%} gain demonstrates that enforcing binary decisions is imperative for overcoming model inertia and recovering fine-grained discriminability.

\begin{table}[t]
\centering
\caption{\textbf{Judge alignment across model scales.} We evaluate Qwen3-VL variants (detailed in Section~\ref{sec:judge_alignment}) and find that larger models align better with human perception. \textit{32B-FP8} yields the best overall performance and is chosen as our judge model.}
\label{tab:judge_acc_elobench}
\resizebox{0.9\linewidth}{!}{
\begin{tabular}{l|cccc}
\toprule
\textbf{Size} & Basic & Reasoning & MultiRef & Overall \\
\midrule
4B   & 61.5 & 65.3 & 56.8 & 60.9 \\
8B   & 64.7 & 66.1 & 57.6 & 63.6 \\
32B   & \textbf{68.2} & \textbf{71.5} & \underline{63.2} & \underline{67.6} \\
32B$^{\mathrm{FP8}}$ & \underline{68.1} & \underline{70.8} & \textbf{66.3} & \textbf{68.0} \\
\bottomrule
\end{tabular}
}
\vspace{-2mm}
\end{table}

\subsubsection{Judge's Alignment with Human Perferences}
\label{sec:judge_alignment}

To evaluate the reliability of automated evaluation, we conduct a scalability analysis using the Qwen3-VL Instruct model family across varying parameter sizes (4B, 8B, and 32B). We assess their alignment with human perception using a subset of preference pairs from~\cite{hu2025multimodal}, filtered to correspond with the three core capabilities of our benchmark. Following the pairwise protocol outlined in Section~\ref{sec:benchmark}, we calculate the accuracy of each model variant in predicting the ground-truth human choice.

The results, detailed in Table~\ref{tab:judge_acc_elobench}, demonstrate a strong positive correlation between model scale and alignment accuracy. While smaller models (4B and 8B) show limited capability, the 32B variants deliver a significant performance leap. Most notably, the \textit{Qwen3-VL-32B Instruct FP8} model achieves the highest overall accuracy of 68.0\%, exhibiting particular robustness in the challenging MultiRef split (66.3\%). Driven by this superior alignment and computationally efficient inference, we designate \textit{Qwen3-VL-32B Instruct FP8} as the default judge for the \ours framework.

\section{Conclusion}

In this work, we expose the reliability bottlenecks of absolute pointwise scoring and propose \ours, a robust pairwise evaluation framework. Our empirical results demonstrate that this methodological shift unlocks state-of-the-art judge accuracy in off-the-shelf open-source VLMs, surprisingly surpassing proprietary counterparts without parameter updates. By achieving superior alignment with human preference, \ours establishes a rigorous, reproducible, and scalable standard for benchmarking the next generation of visual synthesis models.


\section*{Impact Statement}

This paper presents a robust evaluation framework for visual generation, aiming to improve the reliability and reproducibility of automated benchmarks. By demonstrating that open-source VLMs can serve as effective judges via a pairwise paradigm, our work contributes to the democratization of AI research, reducing the community's reliance on costly proprietary models. However, we acknowledge that VLM-based evaluators may inherit biases present in their training data, potentially propagating societal stereotypes into the ranking of generative models. Users of such benchmarks should be aware of these inherent limitations. While better evaluation accelerates the development of generative models, which carries risks of misuse (e.g., deepfakes), we believe that rigorous and aligned measurement is a prerequisite for developing safer and more controllable AI systems.

\section*{Acknowledgement}

The authors would like to thank the artists and other members of FTG for insightful discussions and feedback on the task definition, as well as X. Zhou from Tsinghua University, Zhehao Lin from Nanjing University and Yeyao Ma from Shanghai Jiao Tong University for their contributions to the discussion on inference optimization techniques.


\bibliography{example_paper}

@article{li2024genai,
  title={Genai-bench: Evaluating and improving compositional text-to-visual generation},
  author={Li, Baiqi and Lin, Zhiqiu and Pathak, Deepak and Li, Jiayao and Fei, Yixin and Wu, Kewen and Ling, Tiffany and Xia, Xide and Zhang, Pengchuan and Neubig, Graham and others},
  journal={arXiv preprint arXiv:2406.13743},
  year={2024}
}

@article{luo2025editscore,
  title={Editscore: Unlocking online rl for image editing via high-fidelity reward modeling},
  author={Luo, Xin and Wang, Jiahao and Wu, Chenyuan and Xiao, Shitao and Jiang, Xiyan and Lian, Defu and Zhang, Jiajun and Liu, Dong and others},
  journal={arXiv preprint arXiv:2509.23909},
  year={2025}
}

@article{wu2025editreward,
  title={Editreward: A human-aligned reward model for instruction-guided image editing},
  author={Wu, Keming and Jiang, Sicong and Ku, Max and Nie, Ping and Liu, Minghao and Chen, Wenhu},
  journal={arXiv preprint arXiv:2509.26346},
  year={2025}
}

@inproceedings{ldm,
  title={High-resolution image synthesis with latent diffusion models},
  author={Rombach, Robin and Blattmann, Andreas and Lorenz, Dominik and Esser, Patrick and Ommer, Bj{\"o}rn},
  booktitle={Proceedings of the IEEE/CVF conference on computer vision and pattern recognition},
  pages={10684--10695},
  year={2022}
}

@article{ho2020denoising,
  title={Denoising diffusion probabilistic models},
  author={Ho, Jonathan and Jain, Ajay and Abbeel, Pieter},
  journal={Advances in neural information processing systems},
  volume={33},
  pages={6840--6851},
  year={2020}
}

@article{song2020denoising,
  title={Denoising diffusion implicit models},
  author={Song, Jiaming and Meng, Chenlin and Ermon, Stefano},
  journal={arXiv preprint arXiv:2010.02502},
  year={2020}
}

@article{podell2023sdxl,
  title={Sdxl: Improving latent diffusion models for high-resolution image synthesis},
  author={Podell, Dustin and English, Zion and Lacey, Kyle and Blattmann, Andreas and Dockhorn, Tim and M{\"u}ller, Jonas and Penna, Joe and Rombach, Robin},
  journal={arXiv preprint arXiv:2307.01952},
  year={2023}
}

@article{blattmann2023stable,
  title={Stable video diffusion: Scaling latent video diffusion models to large datasets},
  author={Blattmann, Andreas and Dockhorn, Tim and Kulal, Sumith and Mendelevitch, Daniel and Kilian, Maciej and Lorenz, Dominik and Levi, Yam and English, Zion and Voleti, Vikram and Letts, Adam and others},
  journal={arXiv preprint arXiv:2311.15127},
  year={2023}
}

@inproceedings{esser2024scaling,
  title={Scaling rectified flow transformers for high-resolution image synthesis},
  author={Esser, Patrick and Kulal, Sumith and Blattmann, Andreas and Entezari, Rahim and M{\"u}ller, Jonas and Saini, Harry and Levi, Yam and Lorenz, Dominik and Sauer, Axel and Boesel, Frederic and others},
  booktitle={Forty-first international conference on machine learning},
  year={2024}
}

@misc{labs2025flux1kontextflowmatching,
      title={FLUX.1 Kontext: Flow Matching for In-Context Image Generation and Editing in Latent Space},
      author={Black Forest Labs and Stephen Batifol and Andreas Blattmann and Frederic Boesel and Saksham Consul and Cyril Diagne and Tim Dockhorn and Jack English and Zion English and Patrick Esser and Sumith Kulal and Kyle Lacey and Yam Levi and Cheng Li and Dominik Lorenz and Jonas Müller and Dustin Podell and Robin Rombach and Harry Saini and Axel Sauer and Luke Smith},
      year={2025},
      eprint={2506.15742},
      archivePrefix={arXiv},
      primaryClass={cs.GR},
      url={https://arxiv.org/abs/2506.15742},
}

@article{team2024chameleon,
  title={Chameleon: Mixed-modal early-fusion foundation models},
  author={Team, Chameleon},
  journal={arXiv preprint arXiv:2405.09818},
  year={2024}
}

@article{wang2024emu3,
  title={Emu3: Next-token prediction is all you need},
  author={Wang, Xinlong and Zhang, Xiaosong and Luo, Zhengxiong and Sun, Quan and Cui, Yufeng and Wang, Jinsheng and Zhang, Fan and Wang, Yueze and Li, Zhen and Yu, Qiying and others},
  journal={arXiv preprint arXiv:2409.18869},
  year={2024}
}

@article{pan2025transfer,
  title={Transfer between modalities with metaqueries},
  author={Pan, Xichen and Shukla, Satya Narayan and Singh, Aashu and Zhao, Zhuokai and Mishra, Shlok Kumar and Wang, Jialiang and Xu, Zhiyang and Chen, Jiuhai and Li, Kunpeng and Juefei-Xu, Felix and others},
  journal={arXiv preprint arXiv:2504.06256},
  year={2025}
}

@article{bagel,
  title={Emerging properties in unified multimodal pretraining},
  author={Deng, Chaorui and Zhu, Deyao and Li, Kunchang and Gou, Chenhui and Li, Feng and Wang, Zeyu and Zhong, Shu and Yu, Weihao and Nie, Xiaonan and Song, Ziang and others},
  journal={arXiv preprint arXiv:2505.14683},
  year={2025}
}

@article{xomni,
  title={X-omni: Reinforcement learning makes discrete autoregressive image generative models great again},
  author={Geng, Zigang and Wang, Yibing and Ma, Yeyao and Li, Chen and Rao, Yongming and Gu, Shuyang and Zhong, Zhao and Lu, Qinglin and Hu, Han and Zhang, Xiaosong and others},
  journal={arXiv preprint arXiv:2507.22058},
  year={2025}
}

@misc{google2025gemini25flashimage,
  author = {Google},
  title = {Introducing {Gemini 2.5 Flash Image}, our state-of-the-art image model},
  howpublished = {\url{https://developers.googleblog.com/introducing-gemini-2-5-flash-image/}},
  year = {2025},
  month = {August}
}

@misc{flux-2-2025,
    author={Black Forest Labs},
    title={{FLUX.2: Frontier Visual Intelligence}},
    year={2025},
    howpublished={\url{https://bfl.ai/blog/flux-2}},
}

@article{wu2025qwen,
  title={Qwen-image technical report},
  author={Wu, Chenfei and Li, Jiahao and Zhou, Jingren and Lin, Junyang and Gao, Kaiyuan and Yan, Kun and Yin, Sheng-ming and Bai, Shuai and Xu, Xiao and Chen, Yilei and others},
  journal={arXiv preprint arXiv:2508.02324},
  year={2025}
}

@article{labs2025flux,
  title={FLUX. 1 Kontext: Flow Matching for In-Context Image Generation and Editing in Latent Space},
  author={Labs, Black Forest and Batifol, Stephen and Blattmann, Andreas and Boesel, Frederic and Consul, Saksham and Diagne, Cyril and Dockhorn, Tim and English, Jack and English, Zion and Esser, Patrick and others},
  journal={arXiv preprint arXiv:2506.15742},
  year={2025}
}

@misc{gptimage1,
  author = {OpenAI},
  title = {Introducing 4o Image Generation},
  howpublished = {\url{https://openai.com/index/introducing-4o-image-generation/}},
  year = {2025},
  month = {March}
}

@article{liu2025step1x,
  title={Step1x-edit: A practical framework for general image editing},
  author={Liu, Shiyu and Han, Yucheng and Xing, Peng and Yin, Fukun and Wang, Rui and Cheng, Wei and Liao, Jiaqi and Wang, Yingming and Fu, Honghao and Han, Chunrui and others},
  journal={arXiv preprint arXiv:2504.17761},
  year={2025}
}

@inproceedings{lmarena,
  title={Chatbot arena: An open platform for evaluating llms by human preference},
  author={Chiang, Wei-Lin and Zheng, Lianmin and Sheng, Ying and Angelopoulos, Anastasios Nikolas and Li, Tianle and Li, Dacheng and Zhu, Banghua and Zhang, Hao and Jordan, Michael and Gonzalez, Joseph E and others},
  booktitle={Forty-first International Conference on Machine Learning},
  year={2024}
}

@article{genaiarena,
  title={Genai arena: An open evaluation platform for generative models},
  author={Jiang, Dongfu and Ku, Max and Li, Tianle and Ni, Yuansheng and Sun, Shizhuo and Fan, Rongqi and Chen, Wenhu},
  journal={Advances in Neural Information Processing Systems},
  volume={37},
  pages={79889--79908},
  year={2024}
}

@misc{aaarena,
  author       = {{Artificial Analysis}},
  title        = {Image Arena: Text-to-Image Model Leaderboard},
  year         = {2025},
  howpublished = {\url{https://artificialanalysis.ai/image/arena}}
}

@inproceedings{zhao2025auto,
  title={Auto-arena: Automating llm evaluations with agent peer battles and committee discussions},
  author={Zhao, Ruochen and Zhang, Wenxuan and Chia, Yew Ken and Xu, Weiwen and Zhao, Deli and Bing, Lidong},
  booktitle={Proceedings of the 63rd Annual Meeting of the Association for Computational Linguistics (Volume 1: Long Papers)},
  pages={4440--4463},
  year={2025}
}

@inproceedings{luo2025videoautoarena,
  title={Videoautoarena: An automated arena for evaluating large multimodal models in video analysis through user simulation},
  author={Luo, Ziyang and Wu, Haoning and Li, Dongxu and Ma, Jing and Kankanhalli, Mohan and Li, Junnan},
  booktitle={Proceedings of the Computer Vision and Pattern Recognition Conference},
  pages={8461--8474},
  year={2025}
}

@article{imgedit,
  title={Imgedit: A unified image editing dataset and benchmark},
  author={Ye, Yang and He, Xianyi and Li, Zongjian and Lin, Bin and Yuan, Shenghai and Yan, Zhiyuan and Hou, Bohan and Yuan, Li},
  journal={arXiv preprint arXiv:2505.20275},
  year={2025}
}

@article{han2025unireditbench,
  title={UniREditBench: A Unified Reasoning-based Image Editing Benchmark},
  author={Han, Feng and Wang, Yibin and Li, Chenglin and Liang, Zheming and Wang, Dianyi and Jiao, Yang and Wei, Zhipeng and Gong, Chao and Jin, Cheng and Chen, Jingjing and others},
  journal={arXiv preprint arXiv:2511.01295},
  year={2025}
}

@article{wu2025kris,
  title={KRIS-Bench: Benchmarking Next-Level Intelligent Image Editing Models},
  author={Wu, Yongliang and Li, Zonghui and Hu, Xinting and Ye, Xinyu and Zeng, Xianfang and Yu, Gang and Zhu, Wenbo and Schiele, Bernt and Yang, Ming-Hsuan and Yang, Xu},
  journal={arXiv preprint arXiv:2505.16707},
  year={2025}
}

@article{krippendorff2011computing,
  title={Computing Krippendorff's alpha-reliability},
  author={Krippendorff, Klaus},
  year={2011}
}

@article{heusel2017gans,
  title={Gans trained by a two time-scale update rule converge to a local nash equilibrium},
  author={Heusel, Martin and Ramsauer, Hubert and Unterthiner, Thomas and Nessler, Bernhard and Hochreiter, Sepp},
  journal={Advances in neural information processing systems},
  volume={30},
  year={2017}
}

@article{stein2023exposing,
  title={Exposing flaws of generative model evaluation metrics and their unfair treatment of diffusion models},
  author={Stein, George and Cresswell, Jesse and Hosseinzadeh, Rasa and Sui, Yi and Ross, Brendan and Villecroze, Valentin and Liu, Zhaoyan and Caterini, Anthony L and Taylor, Eric and Loaiza-Ganem, Gabriel},
  journal={Advances in Neural Information Processing Systems},
  volume={36},
  pages={3732--3784},
  year={2023}
}

@inproceedings{hessel2021clipscore,
  title={Clipscore: A reference-free evaluation metric for image captioning},
  author={Hessel, Jack and Holtzman, Ari and Forbes, Maxwell and Le Bras, Ronan and Choi, Yejin},
  booktitle={Proceedings of the 2021 conference on empirical methods in natural language processing},
  pages={7514--7528},
  year={2021}
}

@inproceedings{jayasumana2024rethinking,
  title={Rethinking fid: Towards a better evaluation metric for image generation},
  author={Jayasumana, Sadeep and Ramalingam, Srikumar and Veit, Andreas and Glasner, Daniel and Chakrabarti, Ayan and Kumar, Sanjiv},
  booktitle={Proceedings of the IEEE/CVF Conference on Computer Vision and Pattern Recognition},
  pages={9307--9315},
  year={2024}
}

@inproceedings{kynkaanniemi2023role,
  title={The Role of ImageNet Classes in Fr{\'e}chet Inception Distance},
  author={Kynk{\"a}{\"a}nniemi, Tuomas and Karras, Tero and Aittala, Miika and Aila, Timo and Lehtinen, Jaakko},
  booktitle={ICLR},
  year={2023}
}

@inproceedings{ku2024viescore,
  title={Viescore: Towards explainable metrics for conditional image synthesis evaluation},
  author={Ku, Max and Jiang, Dongfu and Wei, Cong and Yue, Xiang and Chen, Wenhu},
  booktitle={Proceedings of the 62nd Annual Meeting of the Association for Computational Linguistics (Volume 1: Long Papers)},
  pages={12268--12290},
  year={2024}
}

@inproceedings{lin2024evaluating,
  title={Evaluating text-to-visual generation with image-to-text generation},
  author={Lin, Zhiqiu and Pathak, Deepak and Li, Baiqi and Li, Jiayao and Xia, Xide and Neubig, Graham and Zhang, Pengchuan and Ramanan, Deva},
  booktitle={European Conference on Computer Vision},
  pages={366--384},
  year={2024},
  organization={Springer}
}

@inproceedings{chen2024mllm,
  title={Mllm-as-a-judge: Assessing multimodal llm-as-a-judge with vision-language benchmark},
  author={Chen, Dongping and Chen, Ruoxi and Zhang, Shilin and Wang, Yaochen and Liu, Yinuo and Zhou, Huichi and Zhang, Qihui and Wan, Yao and Zhou, Pan and Sun, Lichao},
  booktitle={Forty-first International Conference on Machine Learning},
  year={2024}
}

@article{cao2025hunyuanimage,
  title={Hunyuanimage 3.0 technical report},
  author={Cao, Siyu and Chen, Hangting and Chen, Peng and Cheng, Yiji and Cui, Yutao and Deng, Xinchi and Dong, Ying and Gong, Kipper and Gu, Tianpeng and Gu, Xiusen and others},
  journal={arXiv preprint arXiv:2509.23951},
  year={2025}
}

@article{team2025kimi,
  title={Kimi k2: Open agentic intelligence},
  author={Team, Kimi and Bai, Yifan and Bao, Yiping and Chen, Guanduo and Chen, Jiahao and Chen, Ningxin and Chen, Ruijue and Chen, Yanru and Chen, Yuankun and Chen, Yutian and others},
  journal={arXiv preprint arXiv:2507.20534},
  year={2025}
}

@online{xai2025grok,
  author = {{xAI}},
  title = {Grok 4.1},
  year = {2025},
  month = nov,
  day = {17},
  url = {https://x.ai/news/grok-4-1}
}

@online{google2025gemini3flash,
  author = {{Google}},
  title = {Gemini 3 Flash},
  year = {2025},
  month = dec,
  day = {17},
  url = {https://blog.google/products/gemini/gemini-3-flash/}
}

@article{hu2025multimodal,
  title={Multimodal RewardBench 2: Evaluating Omni Reward Models for Interleaved Text and Image},
  author={Hu, Yushi and Askari-Hemmat, Reyhane and Hall, Melissa and Dinan, Emily and Zettlemoyer, Luke and Ghazvininejad, Marjan},
  journal={arXiv preprint arXiv:2512.16899},
  year={2025}
}

@article{zheng2023judging,
  title={Judging llm-as-a-judge with mt-bench and chatbot arena},
  author={Zheng, Lianmin and Chiang, Wei-Lin and Sheng, Ying and Zhuang, Siyuan and Wu, Zhanghao and Zhuang, Yonghao and Lin, Zi and Li, Zhuohan and Li, Dacheng and Xing, Eric and others},
  journal={Advances in neural information processing systems},
  volume={36},
  pages={46595--46623},
  year={2023}
}

@inproceedings{li2025generation,
  title={From generation to judgment: Opportunities and challenges of llm-as-a-judge},
  author={Li, Dawei and Jiang, Bohan and Huang, Liangjie and Beigi, Alimohammad and Zhao, Chengshuai and Tan, Zhen and Bhattacharjee, Amrita and Jiang, Yuxuan and Chen, Canyu and Wu, Tianhao and others},
  booktitle={Proceedings of the 2025 Conference on Empirical Methods in Natural Language Processing},
  pages={2757--2791},
  year={2025}
}

@article{li2024llms,
  title={Llms-as-judges: a comprehensive survey on llm-based evaluation methods},
  author={Li, Haitao and Dong, Qian and Chen, Junjie and Su, Huixue and Zhou, Yujia and Ai, Qingyao and Ye, Ziyi and Liu, Yiqun},
  journal={arXiv preprint arXiv:2412.05579},
  year={2024}
}

@book{elo1966uscf,
  title={The USCF Rating System: Its Development, Theory, and Applications},
  author={Elo, Arpad E},
  year={1966},
  publisher={United States Chess Federation}
}

@article{bradley1952rank,
  title={Rank analysis of incomplete block designs: I. the method of paired comparisons},
  author={Bradley, Ralph Allan and Terry, Milton E},
  journal={Biometrika},
  volume={39},
  number={3/4},
  pages={324--345},
  year={1952},
  publisher={JSTOR}
}

@incollection{thurstone2017law,
  title={A law of comparative judgment},
  author={Thurstone, Louis L},
  booktitle={Scaling},
  pages={81--92},
  year={2017},
  publisher={Routledge}
}

@inproceedings{radford2021learning,
  title={Learning transferable visual models from natural language supervision},
  author={Radford, Alec and Kim, Jong Wook and Hallacy, Chris and Ramesh, Aditya and Goh, Gabriel and Agarwal, Sandhini and Sastry, Girish and Askell, Amanda and Mishkin, Pamela and Clark, Jack and others},
  booktitle={International conference on machine learning},
  pages={8748--8763},
  year={2021},
  organization={PmLR}
}

@inproceedings{korhonen2012peak,
  title={Peak signal-to-noise ratio revisited: Is simple beautiful?},
  author={Korhonen, Jari and You, Junyong},
  booktitle={2012 Fourth international workshop on quality of multimedia experience},
  pages={37--38},
  year={2012},
  organization={IEEE}
}

@article{wang2004image,
  title={Image quality assessment: from error visibility to structural similarity},
  author={Wang, Zhou and Bovik, Alan C and Sheikh, Hamid R and Simoncelli, Eero P},
  journal={IEEE transactions on image processing},
  volume={13},
  number={4},
  pages={600--612},
  year={2004},
  publisher={IEEE}
}

@inproceedings{brooks2023instructpix2pix,
  title={Instructpix2pix: Learning to follow image editing instructions},
  author={Brooks, Tim and Holynski, Aleksander and Efros, Alexei A},
  booktitle={Proceedings of the IEEE/CVF conference on computer vision and pattern recognition},
  pages={18392--18402},
  year={2023}
}

@inproceedings{sheynin2024emu,
  title={Emu edit: Precise image editing via recognition and generation tasks},
  author={Sheynin, Shelly and Polyak, Adam and Singer, Uriel and Kirstain, Yuval and Zohar, Amit and Ashual, Oron and Parikh, Devi and Taigman, Yaniv},
  booktitle={Proceedings of the IEEE/CVF Conference on Computer Vision and Pattern Recognition},
  pages={8871--8879},
  year={2024}
}

@article{zhang2023magicbrush,
  title={Magicbrush: A manually annotated dataset for instruction-guided image editing},
  author={Zhang, Kai and Mo, Lingbo and Chen, Wenhu and Sun, Huan and Su, Yu},
  journal={Advances in Neural Information Processing Systems},
  volume={36},
  pages={31428--31449},
  year={2023}
}

@article{liu2025inference,
  title={Inference-time scaling for generalist reward modeling},
  author={Liu, Zijun and Wang, Peiyi and Xu, Runxin and Ma, Shirong and Ruan, Chong and Li, Peng and Liu, Yang and Wu, Yu},
  journal={arXiv preprint arXiv:2504.02495},
  year={2025}
}

@article{wang2025unified,
  title={Unified reward model for multimodal understanding and generation},
  author={Wang, Yibin and Zang, Yuhang and Li, Hao and Jin, Cheng and Wang, Jiaqi},
  journal={arXiv preprint arXiv:2503.05236},
  year={2025}
}

@article{Gu2024ASO,
  title={A Survey on LLM-as-a-Judge},
  author={Jiawei Gu and Xuhui Jiang and Zhichao Shi and Hexiang Tan and Xuehao Zhai and Chengjin Xu and Wei Li and Yinghan Shen and Shengjie Ma and Honghao Liu and Yuanzhuo Wang and Jian Guo},
  journal={ArXiv},
  year={2024},
  volume={abs/2411.15594},
}

@article{tan2024blinded,
  title={Blinded by generated contexts: How language models merge generated and retrieved contexts when knowledge conflicts?},
  author={Tan, Hexiang and Sun, Fei and Yang, Wanli and Wang, Yuanzhuo and Cao, Qi and Cheng, Xueqi},
  journal={arXiv preprint arXiv:2401.11911},
  year={2024}
}

@article{ye2024justice,
  title={Justice or prejudice? quantifying biases in llm-as-a-judge},
  author={Ye, Jiayi and Wang, Yanbo and Huang, Yue and Chen, Dongping and Zhang, Qihui and Moniz, Nuno and Gao, Tian and Geyer, Werner and Huang, Chao and Chen, Pin-Yu and others},
  journal={arXiv preprint arXiv:2410.02736},
  year={2024}
}

@article{shi2024judging,
  title={Judging the judges: A systematic study of position bias in llm-as-a-judge},
  author={Shi, Lin and Ma, Chiyu and Liang, Wenhua and Diao, Xingjian and Ma, Weicheng and Vosoughi, Soroush},
  journal={arXiv preprint arXiv:2406.07791},
  year={2024}
}

@inproceedings{thakur2025judging,
  title={Judging the judges: Evaluating alignment and vulnerabilities in llms-as-judges},
  author={Thakur, Aman Singh and Choudhary, Kartik and Ramayapally, Venkat Srinik and Vaidyanathan, Sankaran and Hupkes, Dieuwke},
  booktitle={Proceedings of the Fourth Workshop on Generation, Evaluation and Metrics (GEM$^2$)},
  pages={404--430},
  year={2025}
}

@inproceedings{wang2024large,
  title={Large language models are not fair evaluators},
  author={Wang, Peiyi and Li, Lei and Chen, Liang and Cai, Zefan and Zhu, Dawei and Lin, Binghuai and Cao, Yunbo and Kong, Lingpeng and Liu, Qi and Liu, Tianyu and others},
  booktitle={Proceedings of the 62nd Annual Meeting of the Association for Computational Linguistics (Volume 1: Long Papers)},
  pages={9440--9450},
  year={2024}
}

@article{raina2024llm,
  title={Is llm-as-a-judge robust? investigating universal adversarial attacks on zero-shot llm assessment},
  author={Raina, Vyas and Liusie, Adian and Gales, Mark},
  journal={arXiv preprint arXiv:2402.14016},
  year={2024}
}

@article{zheng2024cheating,
  title={Cheating automatic llm benchmarks: Null models achieve high win rates},
  author={Zheng, Xiaosen and Pang, Tianyu and Du, Chao and Liu, Qian and Jiang, Jing and Lin, Min},
  journal={arXiv preprint arXiv:2410.07137},
  year={2024}
}

@article{zhao2025one,
  title={One token to fool llm-as-a-judge},
  author={Zhao, Yulai and Liu, Haolin and Yu, Dian and Kung, Sunyuan and Chen, Meijia and Mi, Haitao and Yu, Dong},
  journal={arXiv preprint arXiv:2507.08794},
  year={2025}
}

@article{team2025longcat,
  title={LongCat-Image Technical Report},
  author={Team, Meituan LongCat and Ma, Hanghang and Tan, Haoxian and Huang, Jiale and Wu, Junqiang and He, Jun-Yan and Gao, Lishuai and Xiao, Songlin and Wei, Xiaoming and Ma, Xiaoqi and others},
  journal={arXiv preprint arXiv:2512.07584},
  year={2025}
}

@article{xia2025dreamomni2,
  title={Dreamomni2: Multimodal instruction-based editing and generation},
  author={Xia, Bin and Peng, Bohao and Zhang, Yuechen and Huang, Junjia and Liu, Jiyang and Li, Jingyao and Tan, Haoru and Wu, Sitong and Wang, Chengyao and Wang, Yitong and others},
  journal={arXiv preprint arXiv:2510.06679},
  year={2025}
}

@article{oshima2025multibanana,
  title={MultiBanana: A Challenging Benchmark for Multi-Reference Text-to-Image Generation},
  author={Oshima, Yuta and Miyake, Daiki and Matsutani, Kohsei and Iwasawa, Yusuke and Suzuki, Masahiro and Matsuo, Yutaka and Furuta, Hiroki},
  journal={arXiv preprint arXiv:2511.22989},
  year={2025}
}

@article{wu2025omnigen2,
  title={OmniGen2: Exploration to Advanced Multimodal Generation},
  author={Wu, Chenyuan and Zheng, Pengfei and Yan, Ruiran and Xiao, Shitao and Luo, Xin and Wang, Yueze and Li, Wanli and Jiang, Xiyan and Liu, Yexin and Zhou, Junjie and others},
  journal={arXiv preprint arXiv:2506.18871},
  year={2025}
}

@misc{bai2025qwen3vltechnicalreport,
      title={Qwen3-VL Technical Report}, 
      author={Shuai Bai and Yuxuan Cai and Ruizhe Chen and Keqin Chen and Xionghui Chen and Zesen Cheng and Lianghao Deng and Wei Ding and Chang Gao and Chunjiang Ge and Wenbin Ge and Zhifang Guo and Qidong Huang and Jie Huang and Fei Huang and Binyuan Hui and Shutong Jiang and Zhaohai Li and Mingsheng Li and Mei Li and Kaixin Li and Zicheng Lin and Junyang Lin and Xuejing Liu and Jiawei Liu and Chenglong Liu and Yang Liu and Dayiheng Liu and Shixuan Liu and Dunjie Lu and Ruilin Luo and Chenxu Lv and Rui Men and Lingchen Meng and Xuancheng Ren and Xingzhang Ren and Sibo Song and Yuchong Sun and Jun Tang and Jianhong Tu and Jianqiang Wan and Peng Wang and Pengfei Wang and Qiuyue Wang and Yuxuan Wang and Tianbao Xie and Yiheng Xu and Haiyang Xu and Jin Xu and Zhibo Yang and Mingkun Yang and Jianxin Yang and An Yang and Bowen Yu and Fei Zhang and Hang Zhang and Xi Zhang and Bo Zheng and Humen Zhong and Jingren Zhou and Fan Zhou and Jing Zhou and Yuanzhi Zhu and Ke Zhu},
      year={2025},
      eprint={2511.21631},
      archivePrefix={arXiv},
      primaryClass={cs.CV},
      url={https://arxiv.org/abs/2511.21631}, 
}

@misc{vteam2025glm45vglm41vthinkingversatilemultimodal,
      title={GLM-4.5V and GLM-4.1V-Thinking: Towards Versatile Multimodal Reasoning with Scalable Reinforcement Learning},
      author={V Team and Wenyi Hong and Wenmeng Yu and Xiaotao Gu and Guo Wang and Guobing Gan and Haomiao Tang and Jiale Cheng and Ji Qi and Junhui Ji and Lihang Pan and Shuaiqi Duan and Weihan Wang and Yan Wang and Yean Cheng and Zehai He and Zhe Su and Zhen Yang and Ziyang Pan and Aohan Zeng and Baoxu Wang and Bin Chen and Boyan Shi and Changyu Pang and Chenhui Zhang and Da Yin and Fan Yang and Guoqing Chen and Jiazheng Xu and Jiale Zhu and Jiali Chen and Jing Chen and Jinhao Chen and Jinghao Lin and Jinjiang Wang and Junjie Chen and Leqi Lei and Letian Gong and Leyi Pan and Mingdao Liu and Mingde Xu and Mingzhi Zhang and Qinkai Zheng and Sheng Yang and Shi Zhong and Shiyu Huang and Shuyuan Zhao and Siyan Xue and Shangqin Tu and Shengbiao Meng and Tianshu Zhang and Tianwei Luo and Tianxiang Hao and Tianyu Tong and Wenkai Li and Wei Jia and Xiao Liu and Xiaohan Zhang and Xin Lyu and Xinyue Fan and Xuancheng Huang and Yanling Wang and Yadong Xue and Yanfeng Wang and Yanzi Wang and Yifan An and Yifan Du and Yiming Shi and Yiheng Huang and Yilin Niu and Yuan Wang and Yuanchang Yue and Yuchen Li and Yutao Zhang and Yuting Wang and Yu Wang and Yuxuan Zhang and Zhao Xue and Zhenyu Hou and Zhengxiao Du and Zihan Wang and Peng Zhang and Debing Liu and Bin Xu and Juanzi Li and Minlie Huang and Yuxiao Dong and Jie Tang},
      year={2025},
      eprint={2507.01006},
      archivePrefix={arXiv},
      primaryClass={cs.CV},
      url={https://arxiv.org/abs/2507.01006},
}

@article{wang2025internvl3,
  title={Internvl3. 5: Advancing open-source multimodal models in versatility, reasoning, and efficiency},
  author={Wang, Weiyun and Gao, Zhangwei and Gu, Lixin and Pu, Hengjun and Cui, Long and Wei, Xingguang and Liu, Zhaoyang and Jing, Linglin and Ye, Shenglong and Shao, Jie and others},
  journal={arXiv preprint arXiv:2508.18265},
  year={2025}
}

@misc{openai2025gptimage15,
  title = {GPT Image 1.5 Model},
  author = {OpenAI},
  year = {2025},
  howpublished = {\url{https://platform.openai.com/docs/models/gpt-image-1.5}}
}

@article{ariely1998predictably,
  title={Predictably irrational: the hidden forces that shape our decisions},
  author={Ariely, Dan},
  journal={Ebook, Revised and},
  year={1998}
}

@online{blackforestlabs2026flux2klein,
  title   = {{FLUX.2 [klein]: Towards Interactive Visual Intelligence}},
  author  = {{Black Forest Labs}},
  year    = {2026},
  month   = jan,
  day     = {15},
  url     = {https://bfl.ai/models/flux-2-klein}
}

@article{zhao2025envisioning,
  title={Envisioning beyond the pixels: Benchmarking reasoning-informed visual editing},
  author={Zhao, Xiangyu and Zhang, Peiyuan and Tang, Kexian and Zhu, Xiaorong and Li, Hao and Chai, Wenhao and Zhang, Zicheng and Xia, Renqiu and Zhai, Guangtao and Yan, Junchi and others},
  journal={arXiv preprint arXiv:2504.02826},
  year={2025}
}

@article{spearman1961proof,
  title={The proof and measurement of association between two things.},
  author={Spearman, Charles},
  year={1961},
  publisher={Appleton-Century-Crofts}
}

@misc{googledeepmind2025geminiimagepro,
  author = {{Google}},
  title = {Gemini 3 Pro Image (Nano Banana Pro)},
  year = {2025},
  month = nov,
  url = {https://deepmind.google/models/gemini-image/pro/}
}

@misc{yang2025wemmuenhancedbridgingvisionlanguage,
      title={WeMMU: Enhanced Bridging of Vision-Language Models and Diffusion Models via Noisy Query Tokens}, 
      author={Jian Yang and Dacheng Yin and Xiaoxuan He and Yong Li and Fengyun Rao and Jing Lyu and Wei Zhai and Yang Cao and Zheng-Jun Zha},
      year={2025},
      eprint={2512.02536},
      archivePrefix={arXiv},
      primaryClass={cs.CV},
      url={https://arxiv.org/abs/2512.02536}, 
}

@article{li2025uniworld,
  title={Uniworld-v2: Reinforce image editing with diffusion negative-aware finetuning and mllm implicit feedback},
  author={Li, Zongjian and Liu, Zheyuan and Zhang, Qihui and Lin, Bin and Wu, Feize and Yuan, Shenghai and Yan, Zhiyuan and Ye, Yang and Yu, Wangbo and Niu, Yuwei and others},
  journal={arXiv preprint arXiv:2510.16888},
  year={2025}
}

@article{liu2025improving,
  title={Improving video generation with human feedback},
  author={Liu, Jie and Liu, Gongye and Liang, Jiajun and Yuan, Ziyang and Liu, Xiaokun and Zheng, Mingwu and Wu, Xiele and Wang, Qiulin and Xia, Menghan and Wang, Xintao and others},
  journal={arXiv preprint arXiv:2501.13918},
  year={2025}
}

@article{xu2024visionreward,
  title={Visionreward: Fine-grained multi-dimensional human preference learning for image and video generation},
  author={Xu, Jiazheng and Huang, Yu and Cheng, Jiale and Yang, Yuanming and Xu, Jiajun and Wang, Yuan and Duan, Wenbo and Yang, Shen and Jin, Qunlin and Li, Shurun and others},
  journal={arXiv preprint arXiv:2412.21059},
  year={2024}
}

@article{ghosh2023geneval,
  title={Geneval: An object-focused framework for evaluating text-to-image alignment},
  author={Ghosh, Dhruba and Hajishirzi, Hannaneh and Schmidt, Ludwig},
  journal={Advances in Neural Information Processing Systems},
  volume={36},
  pages={52132--52152},
  year={2023}
}

@article{huang2023t2i,
  title={T2i-compbench: A comprehensive benchmark for open-world compositional text-to-image generation},
  author={Huang, Kaiyi and Sun, Kaiyue and Xie, Enze and Li, Zhenguo and Liu, Xihui},
  journal={Advances in Neural Information Processing Systems},
  volume={36},
  pages={78723--78747},
  year={2023}
}

@article{sun2025t2i,
  title={T2i-reasonbench: Benchmarking reasoning-informed text-to-image generation},
  author={Sun, Kaiyue and Fang, Rongyao and Duan, Chengqi and Liu, Xian and Liu, Xihui},
  journal={arXiv preprint arXiv:2508.17472},
  year={2025}
}

@article{niu2025wise,
  title={Wise: A world knowledge-informed semantic evaluation for text-to-image generation},
  author={Niu, Yuwei and Ning, Munan and Zheng, Mengren and Jin, Weiyang and Lin, Bin and Jin, Peng and Liao, Jiaqi and Feng, Chaoran and Ning, Kunpeng and Zhu, Bin and others},
  journal={arXiv preprint arXiv:2503.07265},
  year={2025}
}

@article{liu2023fetv,
  title={Fetv: A benchmark for fine-grained evaluation of open-domain text-to-video generation},
  author={Liu, Yuanxin and Li, Lei and Ren, Shuhuai and Gao, Rundong and Li, Shicheng and Chen, Sishuo and Sun, Xu and Hou, Lu},
  journal={Advances in Neural Information Processing Systems},
  volume={36},
  pages={62352--62387},
  year={2023}
}

@inproceedings{liu2024evalcrafter,
  title={Evalcrafter: Benchmarking and evaluating large video generation models},
  author={Liu, Yaofang and Cun, Xiaodong and Liu, Xuebo and Wang, Xintao and Zhang, Yong and Chen, Haoxin and Liu, Yang and Zeng, Tieyong and Chan, Raymond and Shan, Ying},
  booktitle={Proceedings of the IEEE/CVF Conference on Computer Vision and Pattern Recognition},
  pages={22139--22149},
  year={2024}
}

@inproceedings{huang2024vbench,
  title={Vbench: Comprehensive benchmark suite for video generative models},
  author={Huang, Ziqi and He, Yinan and Yu, Jiashuo and Zhang, Fan and Si, Chenyang and Jiang, Yuming and Zhang, Yuanhan and Wu, Tianxing and Jin, Qingyang and Chanpaisit, Nattapol and others},
  booktitle={Proceedings of the IEEE/CVF Conference on Computer Vision and Pattern Recognition},
  pages={21807--21818},
  year={2024}
}

@article{wang2025pref,
  title={Pref-grpo: Pairwise preference reward-based grpo for stable text-to-image reinforcement learning},
  author={Wang, Yibin and Li, Zhimin and Zang, Yuhang and Zhou, Yujie and Bu, Jiazi and Wang, Chunyu and Lu, Qinglin and Jin, Cheng and Wang, Jiaqi},
  journal={arXiv preprint arXiv:2508.20751},
  year={2025}
}

@inproceedings{ma2025hpsv3,
  title={Hpsv3: Towards wide-spectrum human preference score},
  author={Ma, Yuhang and Wu, Xiaoshi and Sun, Keqiang and Li, Hongsheng},
  booktitle={Proceedings of the IEEE/CVF International Conference on Computer Vision},
  pages={15086--15095},
  year={2025}
}

@article{han2024evalmuse,
  title={Evalmuse-40k: A reliable and fine-grained benchmark with comprehensive human annotations for text-to-image generation model evaluation},
  author={Han, Shuhao and Fan, Haotian and Fu, Jiachen and Li, Liang and Li, Tao and Cui, Junhui and Wang, Yunqiu and Tai, Yang and Sun, Jingwei and Guo, Chunle and others},
  journal={arXiv preprint arXiv:2412.18150},
  year={2024}
}
\bibliographystyle{icml2026}

\newpage
\appendix
\onecolumn

\setcounter{figure}{0}
\setcounter{table}{0}
\renewcommand{\thefigure}{A.\arabic{figure}}
\renewcommand{\thetable}{A.\arabic{table}}

\section{Appendix}

\begin{figure*}[h]
    \centering
    \includegraphics[width=\textwidth]{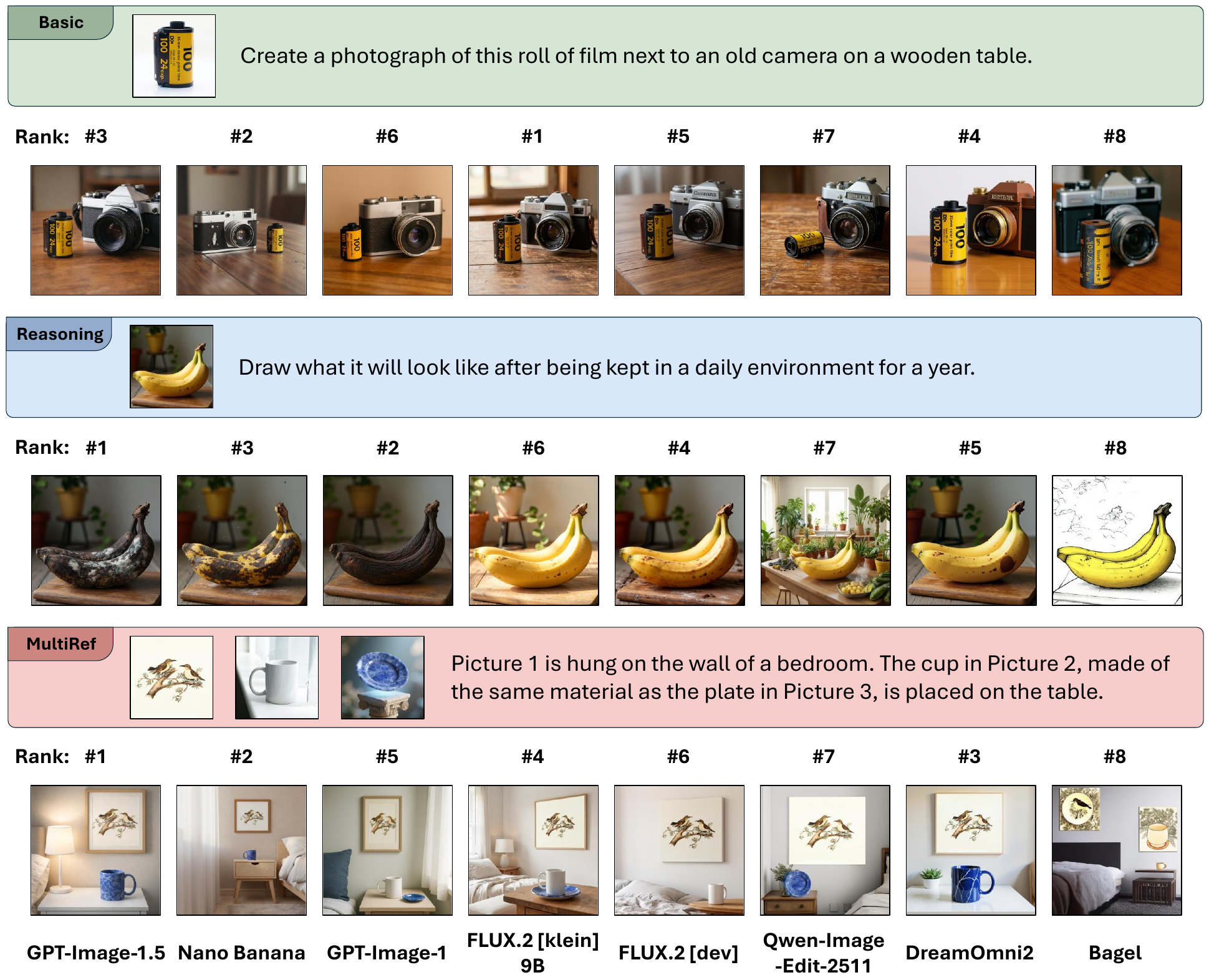}
    \caption{Qualitative comparison of visual generation tasks and generated results from various models in \ours. The benchmark assesses models on three distinct dimensions: Basic, Reasoning (e.g., predicting environmental effects over time), and MultiRef (composing scenes from multiple image conditions). The bottom row displays sample outputs from leading proprietary and open-source models (e.g., GPT-Image-1.5, FLUX.2, Qwen-Image) on a complex multi-reference composition task, highlighting the variance in adherence to spatial and material constraints.}
    \label{fig:samples_of_models}
\end{figure*}

\begin{figure*}[h]
    \centering
    \includegraphics[width=\textwidth]{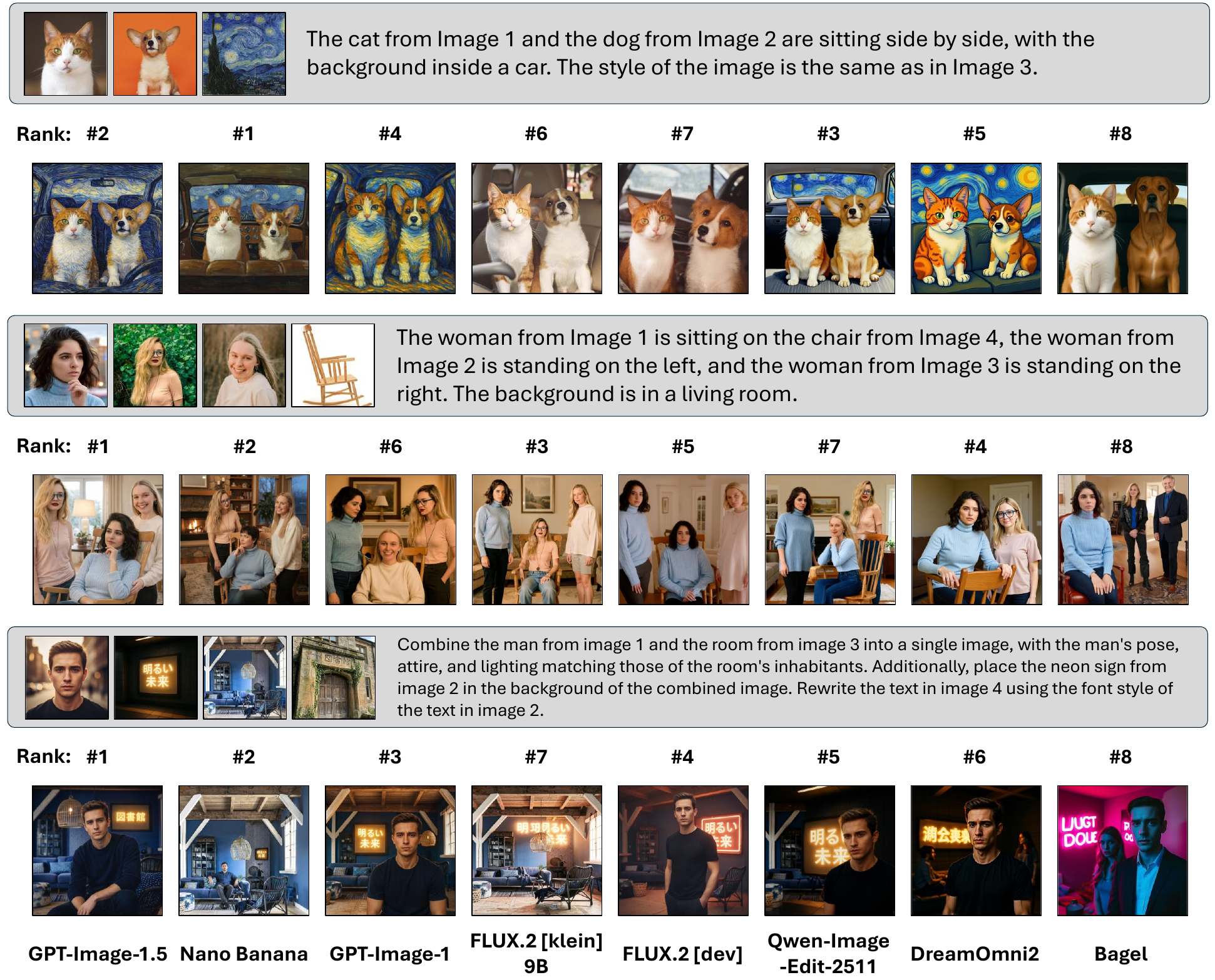}
    \caption{Qualitative comparison of multi-reference generation tasks and generated results from various models in \ours.}
    \label{fig:samples_of_models_multiref}
\end{figure*}

\subsection{Dataset Details}
\label{app:dataset_details}

In Section~\ref{sec:benchmark_composition}, we introduced the three capability dimensions of \ours. Here, we provide the detailed composition and sources of the 6,086 prompts used in our benchmark.

\subsubsection{Source Breakdown}
To ensure the high quality and diversity of our evaluation suite, we aggregated and filtered prompts from the following established benchmarks:

\begin{itemize}
    \item \textbf{Basic Instruction Editing:} Comprises 1,948 prompts sourced from ImgEdit~\cite{imgedit}, GEdit-Bench~\cite{liu2025step1x}, and MMRB2~\cite{hu2025multimodal}. These prompts cover standard editing tasks such as object addition/removal, attribute modification, and background replacement.
    
    \item \textbf{Reasoning-Intensive Editing:} Comprises 1,627 prompts sourced from RISEBench~\cite{zhao2025envisioning} and KRIS-Bench~\cite{wu2025kris}. These prompts are selected to test the model's ability to understand implicit instructions, cultural references, and complex spatial relationships.
    
    \item \textbf{Multi-Reference Composition:} Comprises 2,511 prompts sourced from OmniContext~\cite{wu2025omnigen2}, DreamOmni2Bench~\cite{xia2025dreamomni2}, MultiBanana~\cite{oshima2025multibanana}, and MMRB2~\cite{hu2025multimodal}. This subset represents the most challenging scenarios, requiring the model to understand complicated multimodal instructions and maintain the identity of multiple reference subjects simultaneously.
\end{itemize}

\subsubsection{Qualitative Results}

We provide samples of our benchmark in Figure~\ref{fig:samples_of_models} and Figure~\ref{fig:samples_of_models_multiref}.

\subsection{Pointwise Scoring Causes ``False Tie"}
\label{app:pointwise_analysis}

\begin{figure}[h]
    \centering
    \includegraphics[width=0.9\linewidth]{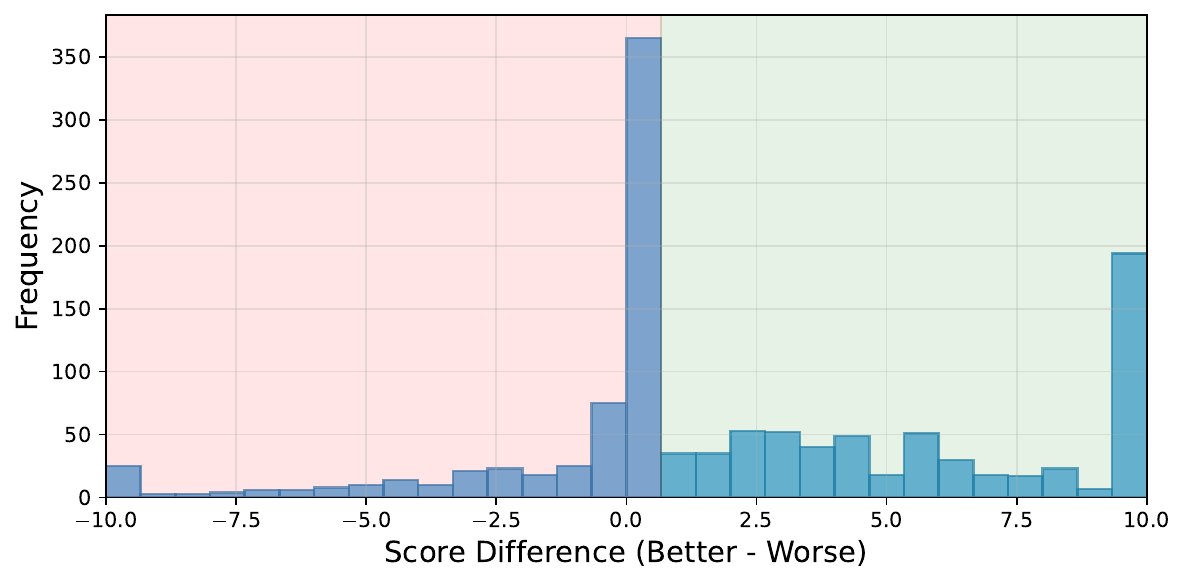}
    \caption{\textbf{Distribution of Score Differences in Pointwise Evaluation.} We visualize the score difference ($\Delta S = S_{\text{better}} - S_{\text{worse}}$) assigned by \textit{Qwen3-VL 8B Instruct} on EditScore-Bench~\cite{luo2025editscore} under the pointwise paradigm. The \textcolor{green!50!black}{green region} ($\Delta S > 0$) denotes correct alignment with human preference, while the \textcolor{red!50!black}{red region} ($\Delta S \leq 0$) indicates contradictory rankings. Notably, the distribution reveals a significant limitation in discriminative power: only 58.3\% of cases are correctly ranked. A substantial portion of comparisons result in ties (23.5\%, spike at $\Delta S = 0$) or direct errors (18.2\%), confirming that absolute pointwise scoring struggles to resolve fine-grained visual differences.}
    \label{fig:pointwise_dist}
\end{figure}

We analyze the distribution of score differences generated by VLM judges. Specifically, we employ \textit{Qwen3-VL 8B Instruct} to evaluate sample pairs from EditScore-Bench~\cite{luo2025editscore} using its pointwise scoring protocol (0-10 scale). For each pair with a known ground-truth human preference, we calculate the score difference $\Delta S = S_{\text{better}} - S_{\text{worse}}$. Ideally, a robust judge should consistently yield $\Delta S > 0$.

However, as illustrated in Figure~\ref{fig:pointwise_dist}, the empirical distribution reveals severe deficiencies in the pointwise scoring mechanism. The model correctly identifies the superior image in only \textbf{58.3\%} of cases (Green Zone). A critical failure mode is the high prevalence of \textbf{ties} (23.5\%), represented by the prominent spike at $\Delta S = 0$. In the context of preference evaluation, a tie is effectively a failure to distinguish quality, rendering the metric useless for ranking closely matched models. Furthermore, \textbf{18.2\%} of the evaluations fall into the negative spectrum (Red Zone), where the judge explicitly favors the inferior image. 

The combined error rate (Ties + Errors) of 41.7\% underscores a fundamental inability of the pointwise paradigm to capture subtle visual nuances. The VLM tends to map distinct visual outputs to the same coarse scalar values, lacking the sensitivity required for high-fidelity image editing assessment. This empirical evidence strongly motivates our transition to the pairwise comparison framework, which forces explicit binary decision-making to resolve these ambiguities.

\subsection{Consistency Analysis Details}
\label{app:consistency}

\subsubsection{Mathematical Formulation}
To compare the internal stability of pointwise and pairwise judges, we unify their outputs into a categorical preference space $\mathcal{L} = \{A \succ B, B \succ A, \text{Tie}\}$. For the pointwise baseline, given scalar scores $S_A^{(k)}$ and $S_B^{(k)}$ from the $k$-th inference run, the derived preference $l_{ij}^{(k)}$ is determined by the sign of the score differential:
\[
l_{ij}^{(k)} = 
\begin{cases} 
A \succ B & \text{if } S_A^{(k)} > S_B^{(k)} \\
B \succ A & \text{if } S_A^{(k)} < S_B^{(k)} \\
\text{Tie} & \text{if } S_A^{(k)} = S_B^{(k)}
\end{cases}
\]
We calculate Krippendorff's Alpha ($\alpha$) based on the observed ($D_o$) and expected ($D_e$) disagreement across $m$ runs:
\begin{equation}
    \alpha = 1 - \frac{D_o}{D_e}, \quad \text{where } D_o = \frac{1}{n} \sum_{c} \sum_{k} o_{ck} \delta(c, k).
\end{equation}
Here, $\delta(c, k)$ is the nominal difference metric, where $\delta(c, k) = 0$ if $c=k$ and $1$ otherwise. This metric strictly measures the stability of the decision boundary; numerical fluctuations in pointwise scores are only penalized if they alter the win/loss outcome. In this context, $\alpha = 1$ signifies perfect determinism, while $\alpha \approx 0$ implies agreement indistinguishable from random chance.

\subsubsection{Experimental Setup}
To ensure a rigorous evaluation, we curated a testbed from two sources:
\begin{itemize}
    \item \textbf{Human-Annotated Datasets:} We utilize the full evaluation sets of GenAI-Bench~\cite{li2024genai} and EditScore-Bench~\cite{luo2025editscore}.
    \item \textbf{Model Matchups:} We use the full prompt sets from ImgEdit~\cite{imgedit} and GEdit-Bench~\cite{liu2025step1x} to generate outputs using the models listed in Table~\ref{tab:elo_more_align}. From these outputs, we randomly sample 500 pairs to constitute the evaluation set.
\end{itemize}
For each pair, we conduct 5 independent inference runs using \textit{Qwen3-VL 8B Instruct} as the judge. To ensure a strictly controlled comparison, all system prompts and generation hyperparameters (e.g., temperature) are kept identical to those used in the main experiments described in Section~\ref{sec:pairwise_accuracy}.

\subsection{Evaluation Criteria and Prompts for VLM Judge}
\label{app:judge_criteria}

In our pairwise scoring mechanism (Sec.~\ref{sec:pairwise_scoring}), the VLM judge is instructed to evaluate image pairs based on four primary standards. These standards ensure that the judgment captures both semantic precision and perceptual fidelity:

\begin{enumerate}
    \item \textbf{Text Faithfulness}: This metric measures the strictness with which the response follows the text editing prompts. It examines the model's ability to accurately implement specific instructions, such as adding distinct objects or altering stylistic attributes.
    \item \textbf{Image Faithfulness}: This criterion focuses on the preservation of the source image's intrinsic characteristics. The edited response should maintain the original's composition, lighting, and background elements, ensuring that modifications do not disrupt the visual continuity.
    \item \textbf{Overall Image Quality}: This standard assesses the output's aesthetic and technical fidelity. Preference is given to responses that minimize visual flaws, such as artifacts or unnatural distortions, thereby ensuring high-quality generation.
    \item \textbf{Text Rendering}: This metric examines the correctness of embedded text. When present, the text is judged on spelling accuracy, readability, and its harmony within the scene; otherwise, it is marked as ``Not Applicable.''
\end{enumerate}

We use the following system prompt for the VLM-as-a judge, following~\cite{hu2025multimodal}.

\begin{promptbox}
    You are an expert in image editing quality analysis and AI evaluation. Your role is to act as an objective judge for comparing two AI-generated image editing responses to the same prompt. You will evaluate which response is better based on a comprehensive rubric specifically designed for image editing tasks.

**Important Guidelines:**
- Be completely impartial and avoid any position biases
- Ensure that the order in which the responses were presented does not influence your decision
- Do not allow the length of the responses to influence your evaluation
- Do not favor certain model names or types
- Be as objective as possible in your assessment
- Focus on image editing specific factors: faithfulness to editing instructions, preservation of input image elements, and overall editing quality

**Understanding the Content Structure:**
- **[ORIGINAL PROMPT TO MODEL:]**: This is the image editing instruction given to both AI models
- **[INPUT IMAGE FROM PROMPT:]**: This is the source image provided to both models for editing
- **[RESPONSE A:]**: The first model's edited image response
- **[RESPONSE B:]**: The second model's edited image response

Your evaluation must be based on a fine-grained rubric that covers the following criteria. For each criterion, you must provide detailed step-by-step reasoning comparing both responses. You will use a 1-6 scoring scale.

**Evaluation Criteria:**
1. **text_faithfulness:** Which response better adheres to the text editing instruction? Consider how well each response follows the specific editing instructions (e.g., adding objects, changing colors, modifying scenes).

2. **image_faithfulness:** Which response better respects and incorporates the key elements of the input image? Consider how well each response preserves important aspects of the original image (composition, lighting, style, background elements) while making the requested changes.

3. **overall_image_quality:** Which response has better general technical and aesthetic quality, with fewer visual artifacts, distortions, or inconsistencies introduced during the editing process?

4. **text_rendering:** If either response contains rendered text, which one has better text quality (spelling, legibility, integration with the image)? If no text is rendered, state "Not Applicable."

**Scoring Rubric:**
- Score 6 (A is significantly better): Response A is significantly superior across most criteria
- Score 5 (A is marginally better): Response A is noticeably better across several criteria
- Score 4 (Unsure or A is negligibly better): Response A is slightly better or roughly equivalent
- Score 3 (Unsure or B is negligibly better): Response B is slightly better or roughly equivalent
- Score 2 (B is marginally better): Response B is noticeably better across several criteria
- Score 1 (B is significantly better): Response B is significantly superior across most criteria

**Confidence Assessment:**
After your evaluation, assess your confidence in this judgment on a scale of 0.0 to 1.0:

**CRITICAL**: Be EXTREMELY conservative with confidence scores. Most comparisons should be in the 0.2-0.5 range.

- **Very High Confidence (0.8-1.0)**: ONLY for absolutely obvious cases where one response is dramatically better across ALL criteria with zero ambiguity. Use this extremely rarely (less than 10% of cases).
- **High Confidence (0.6-0.7)**: Clear differences but some uncertainty remains. Use sparingly (less than 20% of cases).
- **Medium Confidence (0.4-0.5)**: Noticeable differences but significant uncertainty. This should be your DEFAULT range.
- **Low Confidence (0.2-0.3)**: Very close comparison, difficult to distinguish. Responses are roughly equivalent or have conflicting strengths.
- **Very Low Confidence (0.0-0.1)**: Essentially indistinguishable responses or major conflicting strengths.

**IMPORTANT GUIDELINES**:
- DEFAULT to 0.3-0.5 range for most comparisons
- Only use 0.6+ when you are absolutely certain
- Consider: Could reasonable people disagree on this comparison?
- Consider: Are there any strengths in the "worse" response?
- Consider: How obvious would this be to a human evaluator?
- Remember: Quality assessment is inherently subjective

After your reasoning, you will provide a final numerical score, indicate which response is better, and assess your confidence. You must always output your response in the following structured JSON format:

{
    "reasoning": {
        "text_faithfulness": "YOUR REASONING HERE",
        "image_faithfulness": "YOUR REASONING HERE", 
        "overall_image_quality": "YOUR REASONING HERE",
        "text_rendering": "YOUR REASONING HERE",
        "comparison_summary": "YOUR OVERALL COMPARISON SUMMARY HERE"
    },
    "score": <int 1-6>,
    "better_response": "A" or "B",
    "confidence": <float 0.0-1.0>,
    "confidence_rationale": "YOUR CONFIDENCE ASSESSMENT REASONING HERE"
}
\end{promptbox}

\subsection{Pairwise Scoring Implementation Details}
\label{app:pairwise_details}

In Section~\ref{sec:pairwise_scoring}, we introduced the Bi-Directional Consistency Protocol. Here, we elaborate on the specific logic used to resolve pairwise comparisons:

\textbf{1. Forced-Choice Constraint:} 
Unlike prior methods that allow models to output "Tie" directly, we force the VLM to select a winner. This effectively mitigates the \textit{laziness bias}, where models default to a tie to avoid complex reasoning. The judge is prompted to evaluate based on specific multi-dimensional criteria (e.g., semantic alignment, visual fidelity) provided in the system prompt.

\textbf{2. Consistency Check \& Tie Resolution:} 
For every pair of models $A$ and $B$, we conduct two inference runs:
\begin{itemize}
    \item \textbf{Run 1:} Input order $(O_A, O_B)$. Output: $w_1 \in \{A, B\}$.
    \item \textbf{Run 2:} Input order $(O_B, O_A)$. Output: $w_2 \in \{A, B\}$.
\end{itemize}
The final outcome $S_{AB}$ is determined as follows:
\begin{equation}
    S_{AB} = 
    \begin{cases} 
    1 (\text{Win for } A) & \text{if } w_1 = A \text{ and } w_2 = A \\
    0 (\text{Win for } B) & \text{if } w_1 = B \text{ and } w_2 = B \\
    0.5 (\text{Tie}) & \text{if } w_1 \neq w_2 
    \end{cases}
\end{equation}
A tie ($0.5$) implies that the visual difference is either negligible or that the judge is heavily influenced by position bias (e.g., always preferring the first image). By filtering these instances, we ensure high precision in the final rankings.

\subsection{Elo Rating Optimization}
\label{app:elo_math}

While Section~\ref{sec:elo_ranking} outlines the objective function, this section details the optimization procedure.

\textbf{Data Construction:} 
We construct a win-matrix where each entry $W_{ij}$ aggregates the results from the bi-directional protocol. If a comparison results in a tie ($S_{ij}=0.5$), we increment both $W_{ij}$ and $W_{ji}$ by $0.5$.

\textbf{Optimization:} 
Since the negative log-likelihood is convex, we use Logistic Regression to solve for the ratings $\mathbf{R}$. We effectively treat the problem as training a classifier without a bias term, where the input features are one-hot encoded vectors representing the opposing models. We employ the L-BFGS algorithm or SGD to minimize $-\mathcal{L}(\mathbf{R})$.

\begin{figure*}[h]
    \centering
    \includegraphics[width=0.75\textwidth]{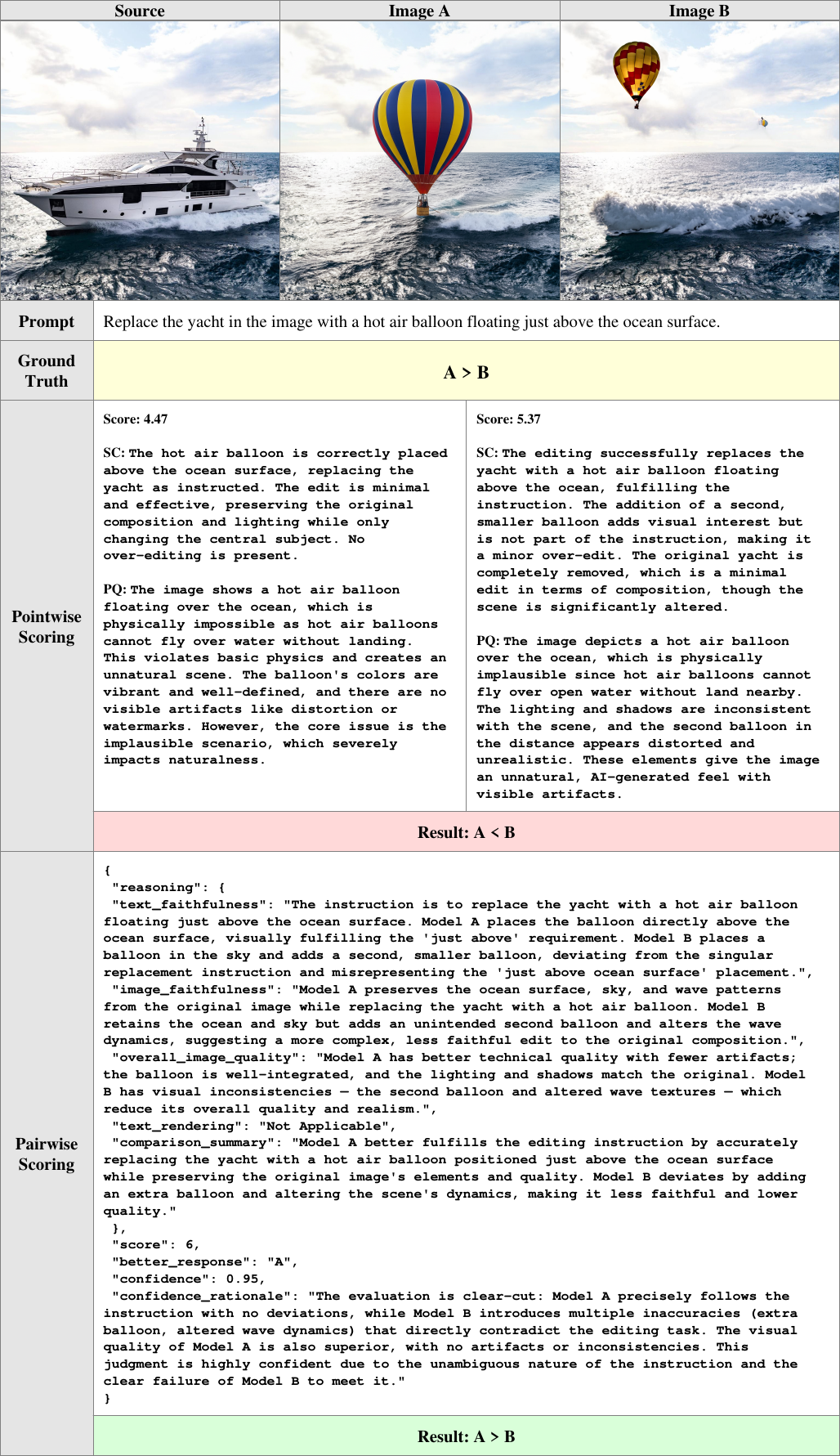}
    \caption{Qualitative comparison of pointwise judgment and pairwise judgment.}
    \label{fig:pairwise_better_1}
\end{figure*}

\begin{figure*}[h]
    \centering
    \includegraphics[width=0.75\textwidth]{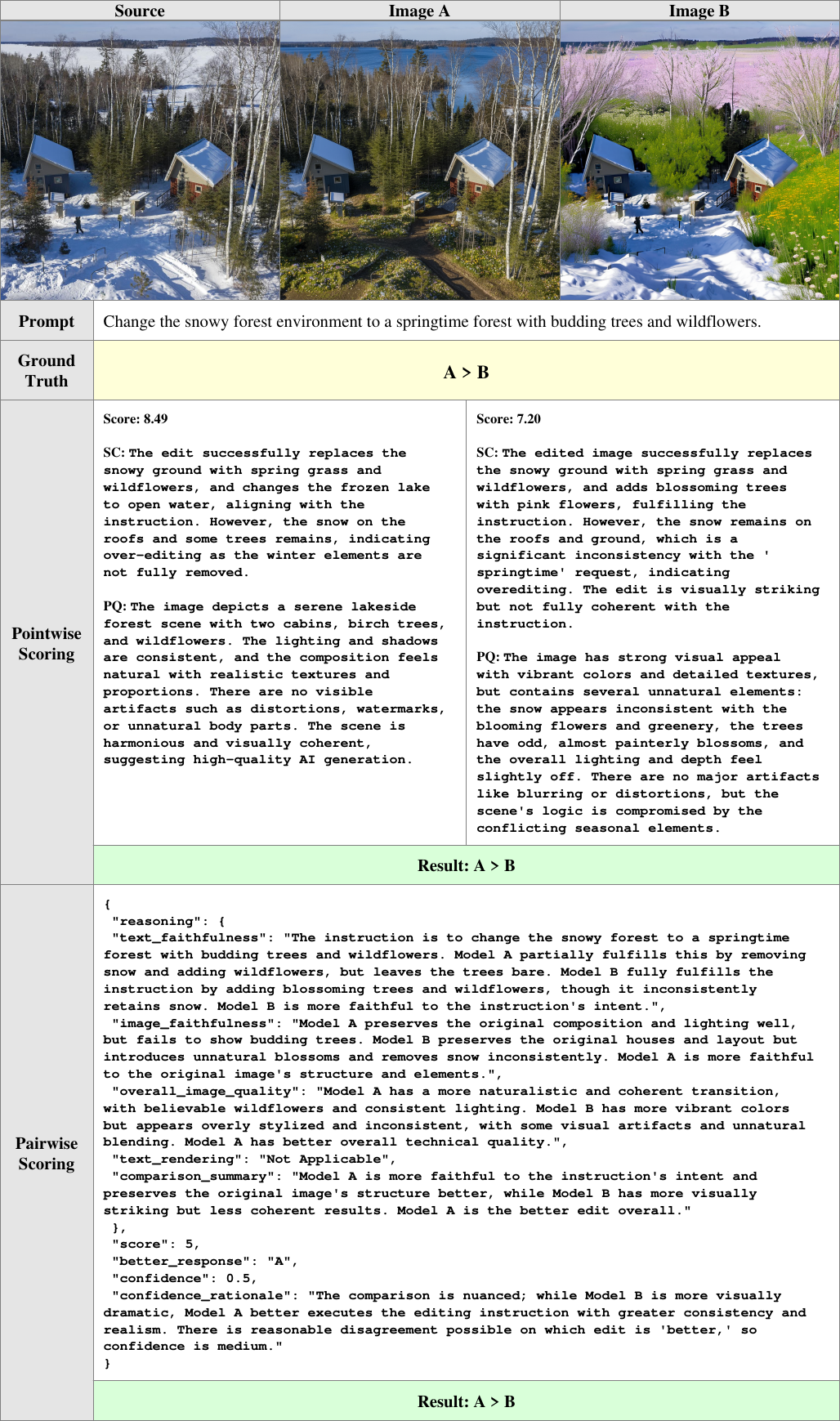}
    \caption{Qualitative comparison of pointwise judgment and pairwise judgment.}
    \label{fig:pairwise_better_2}
\end{figure*}

\begin{figure*}[h]
    \centering
    \includegraphics[width=0.75\textwidth]{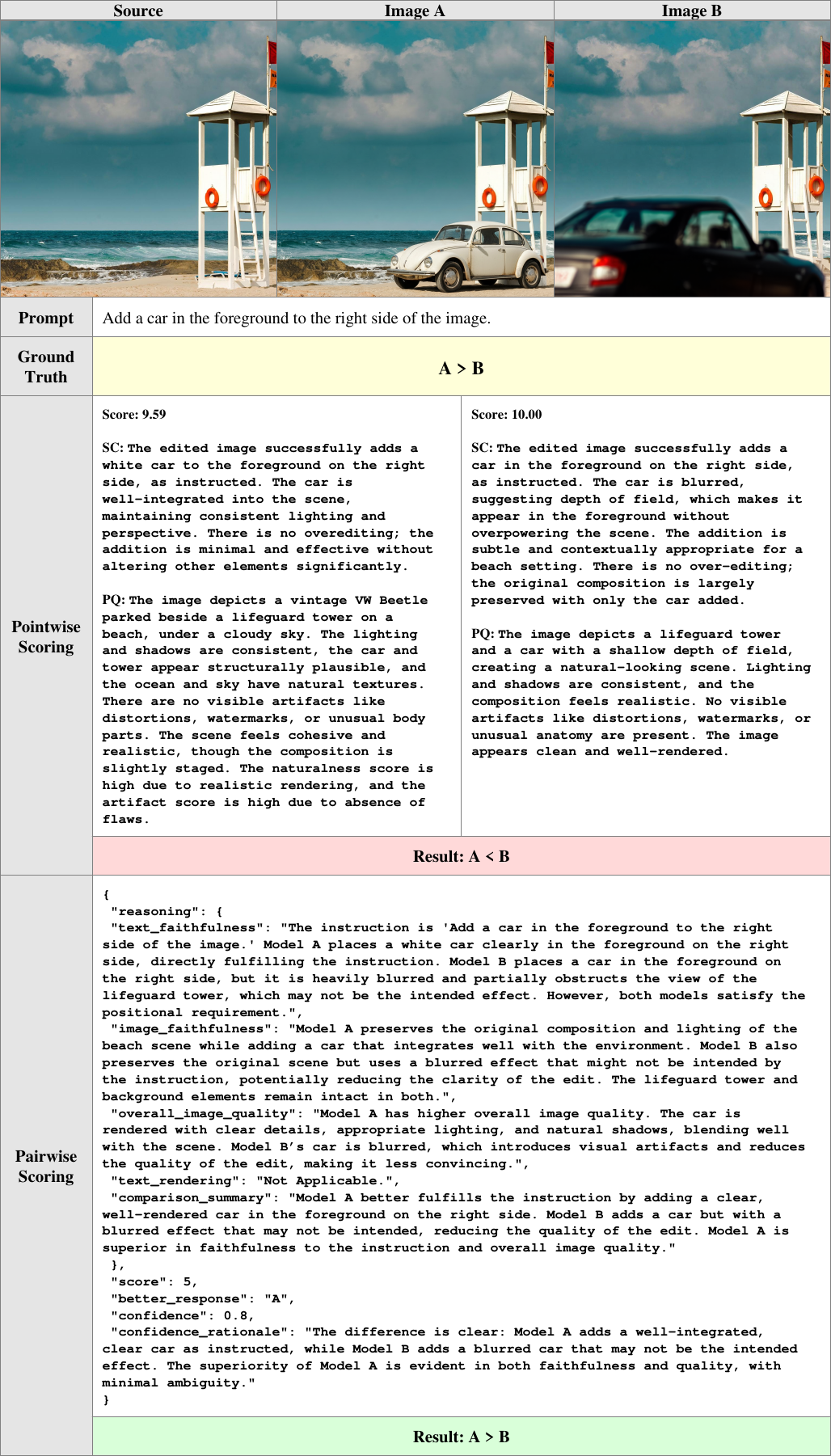}
    \caption{Qualitative comparison of pointwise judgment and pairwise judgment.}
    \label{fig:pairwise_better_3}
\end{figure*}

\end{document}